\documentclass[lettersize,journal]{IEEEtran}
\usepackage{amsmath,amsfonts}
\usepackage{algorithmic}
\usepackage{algorithm}
\usepackage{array}
\usepackage{booktabs}
\usepackage{textcomp}
\usepackage{stfloats}
\usepackage{subfigure}
\usepackage{url}
\usepackage{verbatim}
\usepackage{graphicx}
\usepackage{cite}
\usepackage{xcolor}
\usepackage{caption}
\begin{document}

\title{CoSD: Collaborative Stance Detection with Contrastive Heterogeneous Topic Graph Learning}

\author{
Yinghan~Cheng,~\IEEEmembership{}
Qi~Zhang,~\IEEEmembership{}
Chongyang~Shi,~\IEEEmembership{}
Liang~Xiao,~\IEEEmembership{}
Shufeng~Hao,~\IEEEmembership{}
and~Liang~Hu~\IEEEmembership{}
\thanks{Yinghan Cheng, Chongyang Shi, and Liang Xiao are with the School of Computer Science and Technology, Beijing Institute of Technology, Beijing 100081, China (e-mail: \{chengyinghan,cy\_shi,patrickxiao\}@bit.edu.cn).}
\thanks{Qi Zhang and Liang Hu are with the Department of Computer Science, Tongji University, Shanghai 201804, China (e-mail: \{zhangqi\_cs,lianghu\}@tongji.edu.cn).}
\thanks{Shufeng Hao is with the College of Computer Science and Technology, Taiyuan University of Technology, Taiyuan 030002, China (e-mail: haoshufeng@tyut.edu.cn).}
}

\markboth{Journal of \LaTeX\ Class Files,~Vol.~14, No.~8, August~2021}%
{Shell \MakeLowercase{\textit{et al.}}: A Sample Article Using IEEEtran.cls for IEEE Journals}


\maketitle

\begin{abstract}
Stance detection seeks to identify the viewpoints of individuals either in favor or against a given target or a controversial topic. Current advanced neural models for stance detection typically employ fully parametric softmax classifiers. However, these methods suffer from several limitations, including lack of explainability, insensitivity to the latent data structure, and unimodality, which greatly restrict their performance and applications. To address these challenges, we present a novel \textit{collaborative stance detection} framework called (CoSD) which leverages contrastive heterogeneous topic graph learning to learn topic-aware semantics and collaborative signals among texts, topics, and stance labels for enhancing stance detection. During training, we construct a heterogeneous graph to structurally organize texts and stances through implicit topics via employing latent Dirichlet allocation. We then perform contrastive graph learning to learn heterogeneous node representations, aggregating informative multi-hop collaborative signals via an elaborate Collaboration Propagation Aggregation (CPA) module. During inference, we introduce a hybrid similarity scoring module to enable the comprehensive incorporation of topic-aware semantics and collaborative signals for stance detection. Extensive experiments on two benchmark datasets demonstrate the state-of-the-art detection performance of CoSD, verifying the effectiveness and explainability of our collaborative framework.
\end{abstract}


\section{Introduction}
\IEEEPARstart{S}{tance} detection is a fascinating task that aims to automatically determine the stance or viewpoints expressed towards a particular target, often involving controversial topics or political figures on social media \cite{10.1145/3369026, hardalov-etal-2021-cross}. It allows us to gain insights into public opinion, track sentiment shifts, and analyze the spread of ideas across different communities, applicable to various practical tasks, including but not limited to sarcasm detection \cite{10.1145/3397271.3401183} and fake news detection \cite{10130819,LaoZSCYHM24}. With the exponential growth of user-generated content, understanding the stance of individuals towards various topics has become increasingly significant and challenging \cite{DBLP:conf/icwsm/DarwishSAN20}.

Recently, deep learning models, from Recurrent Neural Networks (RNN) \cite{augenstein-etal-2016-stance} to and Graph Neural Networks (GNN) \cite{liu-etal-2021-enhancing, 10.1145/3442381.3449790}, have achieved the state-of-the-art and gained widespread adoption in stance detection. With these advancements, parametric softmax classifiers have solidified their position as the prevailing regime, which learns a specific set of parameters, i.e., a weight vector and a bias term, for each class~\cite{DBLP:conf/iclr/WangHZL23}. However, these methods suffer from several limitations: 1) \textit{Less explainability}. The parameters in the classification layer are abstract, which is hard to lend to an explanation that humans can process naturally. 2) \textit{Structure insensitivity}. Linear softmax classifiers are typically trained to enhance performance only, which are data structure agnostic and insensitive to inter-stance correlation. 3) \textit{Unimodality}: Only one single weight vector is learned for each stance in a fully parametric manner, essentially assuming unimodality for each class and showing less tolerance of intra-stance variation. As a result, they may not effectively capture subtle differences between/within stances, leading to suboptimal performance of stance detection.

\begin{figure}[!t]
	\centering
        \includegraphics[width=\columnwidth]{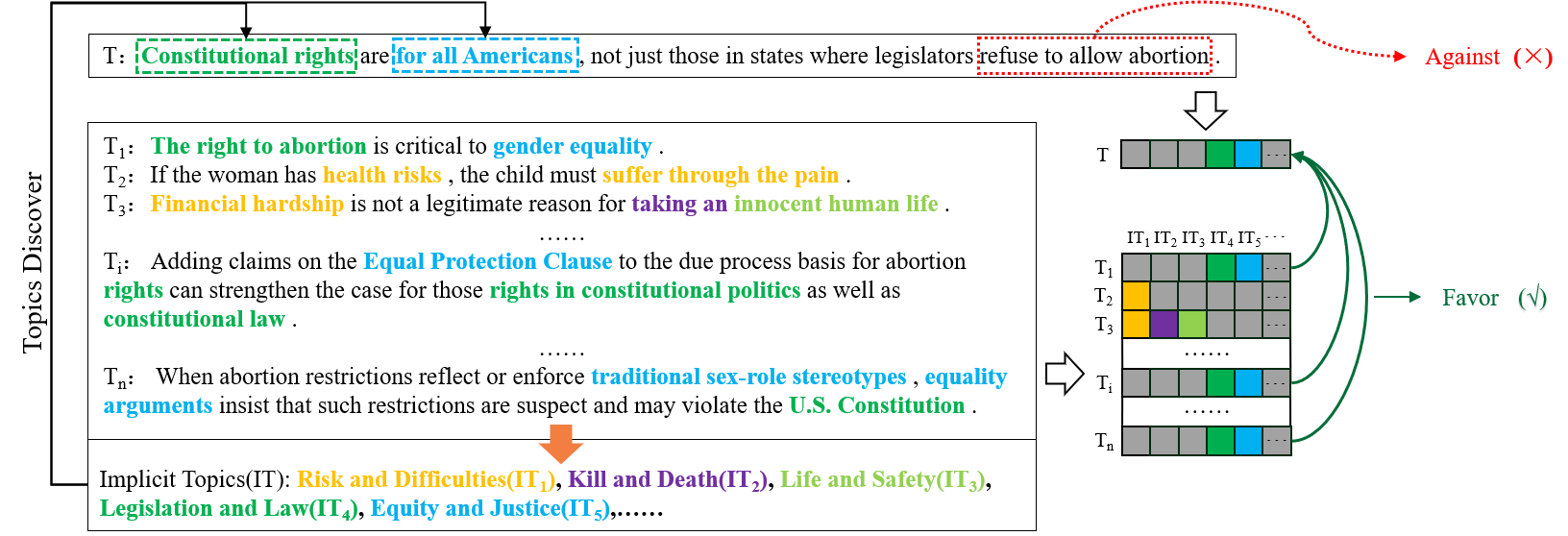}
	\caption{
         An illustrative example demonstrates the impact of collaboration on the test text $T$ and how they contribute to producing the correct result. In the example, the thick brown arrow signifies the generation of implicit topics from the training set. The thin black arrows symbolize the discovery of implicit topics within the test text. The texts $T_1$, $T_i$, and $T_n$ are similar to $T$ in terms of their implicit topic ($IT$) distribution, and they aid $T$ in making accurate judgments, specifically in favor of the target "abortion" in the UKP dataset.}
         \label{fig:problem}
         \vspace{-6mm}
\end{figure}

When expressing their opinions, users often highlight their attitudes towards a specific topic or aspect of the target, while intentionally concealing or downplaying other perspectives or aspects that may be irrelevant~\cite{10.1145/3583780.3615085}. This observation suggests a collaborative phenomenon where texts and their associated targets, sharing similar topics, exhibit inherent correlations. Figure \ref{fig:problem} illustrates this point, where the presence of the phrase 'refuse to allow abortion' (highlighted in red) in the text $T$ tends to indicate a stance 'Against' the goal of 'abortion'. However, if we identify a series of relevant texts that share similar topics with $T$, we can easily arrive at the correct stance of 'Favor'. This is because phrases such as 'Constitutional rights' (highlighted in green) imply the implicit topic of 'Legislation and Law', while phrases like 'for all Americans' (highlighted in blue) imply the implicit topic of 'Equity and Justice'. Obviously, this process in contrast to prevalent parametric softmax classifiers has the potential to address subtle intra/inter-stance differences with a collaborative learning approach to enhance accurate and explainable stance detection.


Additionally, stance detection is a challenging task distinct from sentiment analysis and other text classification tasks, due to the introduction of information-scarce targets. Various efforts including incorporating external knowledge sources~\cite{10.1145/3543507.3583300, DBLP:conf/cncl/WangZYLCL23} or introducing topic modeling~\cite{10.1007/978-3-031-16270-1_18, 10.1145/3543507.3583300} and attention mechanisms\cite{10297287, hanley-durumeric-2023-tata}, have been explored to effectively alleviate the lack of contextual information on targets or explore text-target relationships. However, these approaches generally adopt parametric softmax classifiers as detectors, hardly benefit from the above collaborative phenomenon for exploring the implicit structural relationship between texts and targets. When revisiting topic modeling techniques like latent Dirichlet allocation (LDA), we find implicit topics can serve as a natural bridge between texts, targets, and stance labels. In topic modeling, texts can be represented as a series of implicit topics associated with a specific target, offering the potential of implicit topics to uncover global structural correlations between texts and targets. We can also approximate the distribution of texts within each implicit topic, learning global information for text comprehension. Intuitively, topic modeling not only offers a natural way to correlate texts and targets, but also aligns with collaborative learning to enhance informative target representations and text representations.

In light of the above discussions, we present a novel  \textit{collaborative stance detection} framework (\textbf{CoSD}), which utilizes collaborative signals among texts, topics, and stance labels to enhance stance detection. Specifically, we construct a hierarchical graph correlating texts, targets, and stances through implicit topics extracted via LDA to capture the implicit topics associated with different stances. We then apply contrastive graph learning to learn representations of text, implicit topic, and stance nodes within the graph, leveraging collaborative learning based on the constructed structure. We propose a \textit{Collaboration Propagation Aggregation} (CPA) module. This module enables text representation to aggregate information propagated by implicit topics, similar texts, and higher-order connections in the heterogeneous graph. Additionally, we design a hybrid similarity scoring module that seamlessly integrates BERT \cite{DBLP:conf/naacl/DevlinCLT19} and CPA for inference. The scoring module considers the collaborative information from CPA and the semantic information of BERT, providing comprehensive semantics and topics for stance detection.


The main contributions of this paper are as follows:
\begin{itemize}
\item We propose a novel collaborative stance detection framework CoSD to establish contrastive heterogeneous topic graph learning among texts, targets, and stances. It marks the first attempt to leverage collaborative learning to enhance stance detection effectively.
\item We establish a heterogeneous topic graph to hierarchically correlate texts, targets, and stances using the natural bridge of implicit topics extracted by LDA.
\item We elaborately design a Collaboration Propagation Aggregation (CPA) module to capture informative multi-hop collaborative signals to enhance semantic representation.
\item Extensive experiments on two benchmark datasets show the superior performance of our CoSD over state-of-the-art stance detection baselines, verifying the effectiveness and explainability of our collaborative framework.
\end{itemize}

\section{Related Works}
\subsection{Stance Detection}
Stance detection is aimed at discerning the stance towards a corresponding target. Previous studies primarily concentrate on identifying stances in debates or forums \cite{murakami-raymond-2010-support, somasundaran-wiebe-2010-recognizing, sridhar-etal-2014-collective}. However, since the introduction of the stance detection dataset, a well-known benchmark dataset derived from Twitter and established by the SemEval-2016 competition, there has been a growing interest in conducting stance detection on social media \cite{ALDAYEL2021102597, 10.1145/3369026}. Stance detection is conventionally framed as a three-classification task, with the introduction of the target as a distinguishing factor. Depending on the treatment of the targets, this task can be broadly categorized into two main approaches: supplementary information-based and model-based.

\textbf{Supplementary information-based method:} The contextual information inhabited in the raw texts is often insufficient, necessitating the incorporation of additional information. Representative methods include: 1) External Knowledge Enhancement approaches, which augment the background knowledge of targets or entities by using external knowledge bases\cite{popat-etal-2019-stancy, Kazuaki, 9179107, DBLP:conf/ccks/ZhangYZ021} or involve fine-tuning or providing prompts based on pre-trained language models\cite{10.1145/3477495.3531979}, e.g., GPT \cite{Radford2018ImprovingLU}, RoBERTa \cite{Liu2019RoBERTaAR}, ALBERT \cite{Lan2019ALBERTAL} and ChatGPT. 2) Joint Task Learning approaches, which aid in understanding the text by supplementary tasks \cite{DBLP:journals/fcsc/SunWLZZ19, DBLP:conf/emnlp/LiC19a, chai-etal-2022-improving}. Li et al.\cite{li-etal-2023-new} propose a two-stage framework that first identifies the relevant target in the text and then detects the stance given the target and text.


\textbf{Model-based method:} external-data-based methods may introduce a range of noise, thereby weakening the stance detection task and requiring significant effort to mitigate. Therefore, we prioritize within-data, focusing on mining implicit relational information within the data by models. In earlier work, most of the methods designed based on manually selected features and traditional machine learning \cite{bohler-etal-2016-idi, tutek-etal-2016-takelab}, which are still in use today \cite{GOMEZSUTA2023119046}. More recently, deep learning models have achieved exceptional performance in the detection of stances across various forms. For instance, RNN-based \cite{augenstein-etal-2016-stance} methods leverage the sequential nature of textual data, while CNN-based \cite{wei-etal-2016-pkudblab, 8851965} methods adopt convolutional concepts from computer vision to extract textual features. GNN-based \cite{liu-etal-2021-enhancing, abs-2402-14834} methods focus on the syntactical dependency relation between words. Additionally, popular attention-based methods\cite{10.1145/3544490, 10130819, 10.1145/3543507.3583300} utilize attention mechanisms to capture relations between texts and targets. Wang et al.\cite{Wang2021SolvingSD} applied the target adversarial learning to capture stance-related features shared by all targets and combined target descriptors for learning stance-informative features correlating to specific targets. Li et al.\cite{li-caragea-2023-distilling} uses knowledge distillation over multiple generations for stance detection.


While model-based methods primarily emphasize the target and the direct connection between texts and targets, they often overlook the indirect connection between them and lack in-depth mining of the relationship between texts. A recent study \cite{10.1145/3583780.3615085} models the potential correlation between texts and targets, and clusters similar samples. However, it is essential to consider and leverage both direct and indirect collaborative signals between texts and targets, which is one of our focuses.


\subsection{Collaborative Learning}
In this section, we introduce primary collaborative algorithms in recommendation systems. In the early stages, neighborhood-based methods are predominant and primarily focus on identifying similar users and items, leveraging similarity characterization~\cite{ZhangCSH21,ZhangCSN22}. However, with the advent of matrix factorization techniques, the modeling approach in recommender systems underwent a significant transformation. Matrix factorization entails learning the latent features of users and items by decomposing the user-item rating matrix into a user matrix and an item matrix \cite{5197422, DBLP:conf/uai/RendleFGS09}. While the inner product can force user and item embeddings of an observed interaction close to each other, its linearity makes it insufficient to reveal the complex and nonlinear relationships between users and items \cite{10.1145/3038912.3052569, 10.1145/3038912.3052639}. Following the success of deep learning in areas such as computer vision and natural language processing, Neural Collaborative Filtering (NCF)  \cite{10.1145/3038912.3052569} emerged as a pioneering approach that integrates deep learning into collaborative learning. NCF employs neural networks to capture intricate relationships between users and items, representing users and items using embedded layers. In recent years, the integration of Graph Neural Networks (GNN) into recommendation systems has gained traction, particularly in handling intricate relationships between users and items. NGCF \cite{10.1145/3331184.3331267} stands out as a collaborative learning method based on graph neural networks, designed to effectively capture the neighborhood relationships among users and items. By explicitly encoding collaborative signals and leveraging higher-order connectivity within graphs during the process of learning user and item embeddings, NGCF considers different levels of neighbors, leading to more comprehensive representations.

\subsection{Topic Models}
A topic model is a class of statistical models utilized to uncover the underlying topic structure from text data. Traditional methods include: Latent Dirichlet Allocation (LDA) \cite{10.5555/944919.944937}, Latent Semantic Analysis (LSA)  \cite{DBLP:journals/jasis/DeerwesterDLFH90}, Non-negative Matrix Factorization (NMF), Correlated Topic Models (CTM), etc. LSA employs Singular Value Decomposition (SVD) to reduce the dimension of data, effectively minimizing noise and eliminating redundant information. NMF ensures that all generated matrix components are non-negative. CTM, on the other hand, takes into account the relevance between topics, offering insights into the interrelations among different topics. With the development of deep learning, some topic models based on neural networks have also made significant progress. Pretraining-based Topic Models use pre-trained language models, such as BERT, to model topics to better capture contextual information. Dynamic Topic Models \cite{DBLP:conf/pakdd/ChurchillS22, DBLP:conf/icml/ZhangL22} introduce a time dimension that allows topics to evolve in a collection of documents. Combining Neural networks and Topic modeling, Neural Topic Models \cite{DBLP:conf/acl/PanwarSAK20, DBLP:conf/pkdd/PhamL21} use neural networks to learn text representation and topic representation, which are generally more suitable for large-scale data and deep semantic representation. Despite the ongoing advancements in the field of topic modeling and the emergence of numerous new models and techniques, the interpretability of topic models has been gradually diminishing. Consequently, the significance and efficacy of traditional methods such as LDA cannot be understated. LDA is a generation model that provides a clear explanation of the document generation process, with each topic represented by a distribution of words, facilitating easier interpretation of topics. Moreover, LDA is capable of handling large text corpora and can be trained on extensive document collections, making it a powerful tool for processing massive text datasets. As a result, LDA finds widespread applications in various domains including information retrieval, social media analysis, text summarization, and more.

\section{Method}
\begin{figure*}[htbp]
	\centering
        \includegraphics[width=\textwidth]{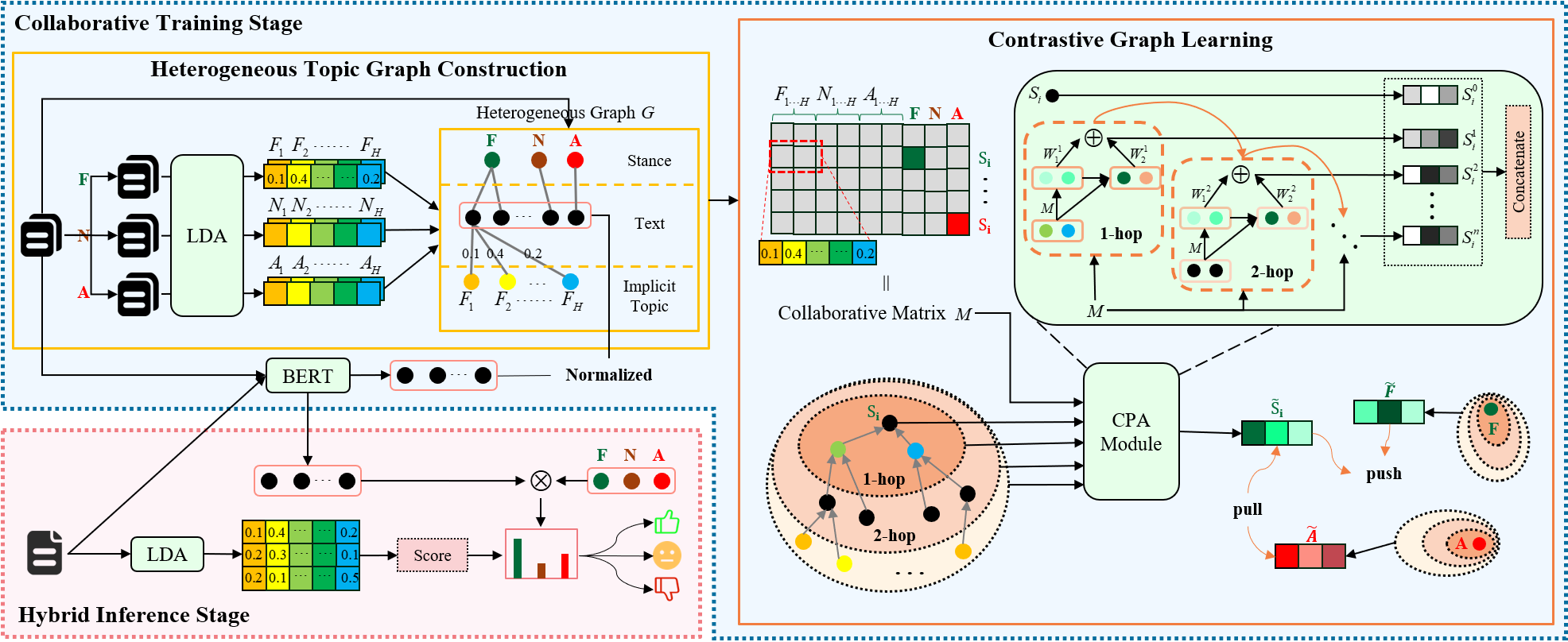}
	\caption{The overall framework structure of our CoSD stance detection method.}
    \label{fig:model}
\end{figure*}

In this section, we present our proposed method, depicted in Figure \ref{fig:model}. Our model comprises three key components: Heterogeneous Topic Graph Construction, Contrastive Graph Training, and Hybrid Inference Stage. Within the Contrastive Graph Training phase, we introduce a novel Collaboration Propagation Aggregation (CPA) module designed to effectively capture collaborative signals.

\subsection{Problem Formulation}
Let denote a text set $S$, a target set $ T = \{ {T_1},{T_2}, \cdots ,{T_m}\} $ and a stance label set $ L = \{ F,N,A\} $, where $m$ is the number of targets, and $\{F,N,A\}$ denotes \{Favor, None, Against\}, respectively. The training set $D_{tr}$ comprises texts paired with their associated targets and corresponding stance labels, denoted as ${D_{tr}} = \{ \left( {{s_i},{t_i},{y_i}} \right)\} _{i = 1}^{{n_{tr}}}$. Meanwhile, the testing set $D_{te}$ is expressed as ${D_{te}} = \{ \left( {{s_i},{t_i}} \right)\} _{i = 1}^{{n_{te}}}$, where $n_{tr}$ and $n_{te}$ represent the size of training and testing sets, respectively. Stance detection aims to discern the stance label $y_i$ expressed by $s_i$ towards $t_i$.

\subsection{Heterogeneous Topic Graph Construction} \label{HTG Module}
In the Collaborative Training Stage, texts corresponding to a target $T_i$ are categorized into favor, none, and against subsets based on their stance labels. Then we employ LDA \cite{10.5555/944919.944937} to generate $H$ implicit topics for each subset and derive the implicit topic distribution for each text, which signifies the probability of belonging to each implicit topic. The implicit topic set $IT$ and the implicit topic distribution $Di{s_i}$ for each text ${s_i}$ are denoted as:
\begin{equation} \label{con:equation1}
\begin{aligned}
  IT &= \{ {F_1}, \cdots ,{F_H},{N_1}, \cdots ,{N_H},{A_1}, \cdots ,{A_H}\} \\
  Di{s_i} &= \{ {f_{i1}}, \cdots ,{f_{iH}},{n_{i1}}, \cdots ,{n_{iH}},{a_{i1}}, \cdots ,{a_{iH}}\}
\end{aligned}
\end{equation}
Each element in $Di{s_i}$ represents the probability of ${s_i}$ belonging to the corresponding implicit topic in $IT$. Utilizing the obtained implicit topics set $IT$, we construct a Heterogeneous Topic Graph $G$ among the implicit topics, texts, and stance labels.  For the edges between the texts and the implicit topics $IT$, we employ the distribution $Di{s_i}$ as the edge weight. Formally, the adjacent matrix $\mathbf{M}_1$ of texts-implicit topics graph is computed as follows:
\begin{equation}
  \mathbf{M}_1 = \{Di{s_i}\}_{i=1}^{n_{tr}}
\end{equation}
Between the texts and the stance labels $L$, we assign the weight of the edge as either 0 or 1. Formally, the adjacent matrix $\mathbf{M}_2$ of texts-stance labels graph is defined as follows:
\begin{equation}
  {\mathbf{M}_2}[i,j] = \left\{ 
  \begin{array}{l}
  1 \qquad if \quad {y_i} = {Y_j}\\
  0 \qquad else
  \end{array} 
  \right.
\end{equation}
where ${y_i}$ represents the stance label corresponding to the text ${s_i}$, and ${Y_j}$ represents the $j$-th stance label in $L$. For each node on the heterogeneous topic graph $G$, the implicit topic nodes are randomly initialized, while for text nodes and stance label nodes, we utilize BERT to obtain their initial representation:
\begin{equation}
  \begin{aligned}
  \label{con:equation4}
  {\mathbf{U}} &= \{ {\mathbf{u}_{{F_1}}}, \cdots ,{\mathbf{u}_{{F_H}}},{\mathbf{u}_{{N_1}}}, \cdots ,{\mathbf{u}_{{N_H}}},{\mathbf{u}_{{A_1}}}, \cdots ,{\mathbf{u}_{{A_H}}}\} \\
  \mathbf{V} &= \{\mathbf{v}_i\}_{i=1}^{n_{tr}}=\{\operatorname{BERT}(s_i)\}_{i=1}^{n_{tr}}\\
  {\mathbf{Z}} &= \{ {\mathbf{z}_F},{\mathbf{z}_N},{\mathbf{z}_A}\}  = \operatorname{BERT}(L)
  \end{aligned}
\end{equation}
where $\mathbf{U}$ denotes the set of representations for implicit topics, $\mathbf{V}$ represents the set of texts' representations, and $\mathbf{Z}$ denotes the set of representations for stance labels.

\subsection{Contrastive Graph Training}
Per Equation \ref{con:equation4}, we construct an embedding representation table as follow:
\begin{equation} \label{con:equation5}
\begin{aligned}
  \mathbf{E} &= [{\mathbf{V}},\mathbf{U},\mathbf{Z}] \\
  &= [\underbrace {{\mathbf{v}_{{1}}}, \cdots ,{\mathbf{v}_{{{n_{tr}}}}}}_{text},\underbrace {{\mathbf{u}_{{F_1}}}, \cdots ,{\mathbf{u}_{{A_H}}}}_{implicit\;topic},\underbrace {{\mathbf{z}_F},{\mathbf{z}_N},{\mathbf{z}_A}}_{stance\;label}]
\end{aligned}
\end{equation}
where we denote $\mathbf{E}=\{\mathbf{e}_i\}_{i=1}^{|\mathbf{E}|}$ as a parameter matrix. This embedding representation table serves as an initial state, subject to optimization in an end-to-end fashion. Throughout the training process, we optimize the embeddings within this table by propagating and aggregating them across the heterogeneous topic graph $G$. To accomplish this, we design a Collaboration propagation aggregation module. Finally, a contrastive loss was employed to refine the training process of the model by constructing pairs of positive and negative samples. This enhances the embeddings for evaluation, as Contrastive Graph Training explicitly injects collaborative signals between texts, implicit topics, and stance labels into the embeddings.

\noindent\textbf{Collaborative Matrix Generation}. We create a collaborative matrix, which serves as input for the subsequent Collaboration Propagation Aggregation Module to optimize the embeddings. The collaborative matrix is defined as:
\begin{equation} \label{con:equation6}
  \mathbf{M} = [\mathbf{M}_1,\mathbf{M}_2]
\end{equation}

\subsection{Collaboration Propagation Aggregation}  \label{CPA Module}
Here, we introduce the Collaboration Propagation Aggregation (CPA) Module, which expands upon the messaging architecture of Graph Neural Networks (GNNs). This module is designed to capture collaborative signals along the graph structure and refine node embeddings. It facilitates the propagation and aggregation of information from multi-hop nodes on the graph, thus fostering collaboration among nodes.

\subsubsection{\textbf{One-Hop Propagation Aggregation}}
First of all, we introduce how the one-hop collaboration information is propagated and aggregated. For each node $\mathbf{e}$ in the embedding representation table $\mathbf{E}$, we define the information propagated from node $\mathbf{e}_i$ in its one-hop node set $\mathcal{E}_1$ as follow:
\begin{equation} \label{con:equation7}
  {I_{{\mathbf{e}_i}}} = \frac{1}{{\sqrt {\left| {{N_e}} \right|\left| {{N_{{e_i}}}} \right|} }}\left( {{\mathbf{W}_1}{\mathbf{e}_i} + {\mathbf{W}_2}\left( {\mathbf{e} \odot {\mathbf{e}_i}} \right)} \right), \quad \mathbf{e}_i\in\mathcal{E}_1
\end{equation}
where ${\mathbf{W}_1}$, ${\mathbf{W}_2}$ represent trainable parameter matrices used to distill valuable information for propagation, with $\odot$ denoting the element-wise product. Here we add the interaction between $\mathbf{e}$ and $\mathbf{e}_i$ through ${\mathbf{e} \odot {\mathbf{e}_i}}$ to capture additional insights from similar nodes and emphasize collaborative phenomena. Following the Graph Convolutional Network (GCN), the graph Laplacian normalization term ${1}/{{\sqrt {\left| {{N_e}} \right|\left| {{N_{{e_i}}}} \right|} }}$, where ${N_e}$ and ${N_{{e_i}}}$ represent the degree of the node $\mathbf{e}$ and node $\mathbf{e}_i$, is utilized as a discount factor to account for the diminishing impact of information propagation with increasing path length.

Then we aggregate the information propagated from $\mathbf{e}$'s one-hop nodes along with its own information to optimize $\mathbf{e}$'s embedding representation. The one-hop propagation aggregation embedding representation of $\mathbf{e}$ is defined as:
\begin{equation}
  {\mathbf{e}^1} = {\mathop{\rm LReLU}\nolimits} \left( {{\mathbf{W}_1}\mathbf{e} + \sum\limits_{{\mathbf{e}_i}} {{I_{{\mathbf{e}_i}}}} } \right), \quad \mathbf{e}_i\in\mathcal{E}_1
\end{equation}
where ${\mathbf{W}_1}$ is the parameter matrices used in Equation \ref{con:equation7}, and here we employ the activation function ${\rm LReLU}$ for its ability to encode both positive and small negative signals.

\subsubsection{\textbf{Multi-Hop Propagation Aggregation}}
Upon the one-hop propagation aggregation, we extend it to multi-hop propagation aggregation to aggregate multi-hop collaboration information. For $\mathbf{e}$, the recursive formula is given by:
\begin{equation} \label{con:equation9}
  \begin{aligned}
  {\mathbf{e}^l} &= {\mathop{\rm LReLU}\nolimits} \left( {\mathbf{W}_1^l{\mathbf{e}^{l - 1}} + \sum\limits_{{\mathbf{e}_i}} {I_{{\mathbf{e}_i}}^l} } \right), \quad \mathbf{e}_i\in\mathcal{E}_l \\
  I_{{\mathbf{e}_i}}^l &= \frac{1}{{\sqrt {\left| {{N_e}} \right|\left| {{N_{{e_i}}}} \right|} }}\left( {\mathbf{W}_1^l\mathbf{e}_i^{l - 1} + \mathbf{W}_2^l\left( {{\mathbf{e}^{l - 1}} \odot \mathbf{e}_i^{l - 1}} \right)} \right)
  \end{aligned}
\end{equation}
where ${\mathbf{W}_1^l}$ and ${\mathbf{W}_2^l}$ denote trainable parameter matrices utilized for transforming between $(l-1)$-hop and $l$-hop nodes respectively, $\mathcal{E}_l$ denotes the $l$-hop node set of $\mathbf{e}$, ${\mathbf{e}^l}$ signifies the $l$-hop propagation aggregation embedding representation of $\mathbf{e}$, and ${I_{{\mathbf{e}_i}}^l}$ indicates the information propagated to $\mathbf{e}$ from one of its $l$-hop node ${\mathbf{e}_i}$. Consequently, the multi-hop collaboration information can be incorporated into node representations.

The matrix representation of Equation \ref{con:equation9} is expressed as:
\begin{equation} \label{con:equation10}
  \begin{aligned}
  {\mathbf{E}^l} &= {\mathop{\rm LReLU}\nolimits} \left( {\mathbf{E}^{l - 1}}{\mathbf{W}_1^l + \mathbf{L}{\mathbf{E}^{l - 1}}\mathbf{W}_1^l + {\mathbf{E}^{l - 1}}\mathbf{W}_2^l \odot \mathbf{L}{\mathbf{E}^{l - 1}}} \right) \\
  &= {\mathop{\rm LReLU}\nolimits} \left( \left( {{\mathop{ \mathbf{I}}\nolimits}  + \mathbf{L}} \right){{\mathbf{E}^{l - 1}}\mathbf{W}_1^l + {\mathbf{E}^{l - 1}}\mathbf{W}_2^l \odot \mathbf{L}{\mathbf{E}^{l - 1}}} \right) 
  \end{aligned} 
\end{equation}
where ${\mathbf{E}^l}$ represents the $l$-hop propagation aggregation embedding representation of all nodes $\mathbf{e}$ with $l$-hop nodes' collaboration information fusion in, ${\mathbf{E}^0}$ denotes $\mathbf{E}$ as mentioned in Equation \ref{con:equation5}, $\mathbf{I}$ signifies the identity matrix, and $\mathbf{L}$ denotes the Laplacian matrix for the adjusted heterogeneous topic graph $G$, defined as follow:
\begin{equation}
    \mathbf{L} = {\mathbf{D}^{ - 1/2}}\left[ \begin{array}{l}
    \mathbf{0}\\
    {\mathbf{M}^T}
    \end{array} \right.\left. \begin{array}{l}
    \mathbf{M}\\
    \mathbf{0}
    \end{array} \right]{\mathbf{D}^{ - 1/2}}
\end{equation}
where $\mathbf{M}$ represents the collaboration matrix generated in Equation \ref{con:equation6}, $\mathbf{0}$ denotes all-zero matrix and $\mathbf{D}$ signifies the diagonal degree matrix.

Using Equation \ref{con:equation10}, we derive $l$ representations of node $\mathbf{e}$, each aggregating the collaboration information from the $l$-hop nodes of $\mathbf{e}$. These representations are subsequently concatenated to obtain the final representation ${\tilde{\mathbf{e}}}$ of node $\mathbf{e}$:
\begin{equation} \label{con:equation12}
  {\tilde{\mathbf{e}}} = \left[ {{\mathbf{e}^0},{\mathbf{e}^1}, \cdots ,{\mathbf{e}^l}} \right]
\end{equation}
where $\mathbf{e}^0$ is obtained from the parameter matrix in Equation \ref{con:equation5}. Leveraging the CPA module, we iteratively refine the node embedding representation, leading to an optimized representation that effectively integrates multi-hop collaboration information.

\subsection{Training Objectives}
The primary goal of CPA is to capture the collaborative relationship between nodes. However, nodes inherently possess their own semantic expressions. To augment the semantic information of the texts, we leverage BERT to acquire representations for both the text $s_i$ and its associated target $t_i$. Subsequently, the attention mechanism is utilized to derive the semantic representation of the text, to emphasize the semantic information pertinent to the target within the text:
\begin{equation}  \label{con:equation13}
\begin{array}{l}
    {\mathbf{e}_{{s_i}}} = \operatorname{BERT}({s_i}), {\mathbf{e}_{{t_i}}} = \operatorname{BERT}({t_i}) \\
    \mathbf{e}_{{s_i}}^{sem} = {\mathop{\rm Attention}\nolimits} \left( {{\mathbf{e}_{{t_i}}},{\mathbf{e}_{{s_i}}},{\mathbf{e}_{{s_i}}}} \right)
\end{array}
\end{equation}
where Attention represents the Dot-Product Attention. Ultimately, we train the entire framework with a combination of contrastive loss and normalized cosine similarity loss:
\begin{equation}
\begin{array}{l}
    \mathcal{L}{s^{con}} =  - {\mathop{\rm In}\nolimits} \sigma \left( {{{\tilde {\mathbf{v}}}_i}^T{{\tilde {\mathbf{z}}}_{y_i}} - {{\tilde {\mathbf{v}}}_i}^T{{\tilde {\mathbf{z}}}_{L - {y_i}}}} \right)\\
    \mathcal{L}{s^{cos}} = \operatorname{Norm}\left( {\operatorname{cos}(e_{{s_i}}^{sem},{\mathbf{v}_i})} \right)\\
    \mathcal{L} = \mathcal{L}{s^{con}} + \mathcal{L}{s^{cos}}
\end{array}
\end{equation}
where ${\tilde {\mathbf{v}}}_i$ and ${\tilde {\mathbf{z}}}_i$ are derived from Equation \ref{con:equation12}, obtained through the representation ${\mathbf{v}}_i$ and ${\mathbf{z}}_i$ of the node $s_i$ and $y_i$ via the CPA Module. $y_i$ denotes the stance label of the ${s_i}$, $L$ represents the stance label set. $\operatorname{cos}$ denotes the cosine similarity loss, $e_{{s_i}}^{sem}$ is obtained from Equation \ref{con:equation13}, and $\mathbf{v}_i$ comes from the parameter matrix in Equation \ref{con:equation5}.

\subsection{Hybrid Inference Stage}
We characterize the similarity between texts, implicit topics, and stance labels by formulating a score, which serves as a crucial component for the stance detection task during the inference stage.

\subsubsection{\textbf{Semantic Score}}
We calculate the inner product of $\mathbf{e}_{{s_i}}^{sem}$ and the stance label representation to delineate the stance tendency of the text's semantic representation:
\begin{equation}
    score_{{s_i}}^{sem} = \mathbf{e}_{{s_i}}^{sem}{\mathbf{Z}^T}
\end{equation}
where $\mathbf{Z}$ denotes the stance label embedding representation obtained from the parameter matrix in Equation \ref{con:equation5}.

\subsubsection{\textbf{Distributed Score}}
For text ${s_i}$, as per the Equation \ref{con:equation1}, we obtain the implicit topic distribution $Di{s_i}$, from which we derive the distribution representation of the text:
\begin{equation}
    e_{{s_i}}^{dis} = \left[ {Di{s_i}} \right]{ \mathbf{U}^T}
\end{equation}
where $\mathbf{U}$ denotes the implicit topic embedding representation obtained from the parameter matrix in Equation \ref{con:equation5}. Subsequently, we calculate the score of the distribution representation against the implicit topic representation to characterize the stance tendency of the text's distributed representation. Initially, we enter $e_{{s_i}}^{dis}$ and $\mathbf{U}$ into the CPA module in section \ref{CPA Module} to obtain $\tilde e_{_{{s_i}}}^{dis}$ and ${\tilde {\mathbf{U}}}$. However, unlike before, the recursive formula in Equation \ref{con:equation9} is modified to:
\begin{equation}
    {\mathbf{e}^l} = {\mathop{\rm LReLU}\nolimits} \left( {\mathbf{W}_1^l{\mathbf{e}^{l - 1}} + \mathbf{W}_2^l{\mathbf{e}^{l - 1}}} \right)
\end{equation}
Next, we obtain the distributed score through the following:
\begin{equation}
\begin{aligned}
    &{\tilde {\mathbf{U}}} = \{ {{\tilde {\mathbf{u}}}_{{F_1}}}, \cdots ,{{\tilde {\mathbf{u}}}_{{F_H}}},{{\tilde {\mathbf{u}}}_{{N_1}}}, \cdots ,{{\tilde {\mathbf{u}}}_{{N_H}}},{{\tilde {\mathbf{u}}}_{{A_1}}}, \cdots ,{{\tilde {\mathbf{u}}}_{{A_H}}}\} \\
    &score_{{s_i}}^{dis} = \left[ {\max \left( {\tilde e_{_{{s_i}}}^{dis}{{\tilde {\mathbf{U}}}_i}} \right)} \right], \quad i \in L
\end{aligned}
\end{equation}
where ${\tilde {\mathbf{U}}}_F=\{ {{\tilde {\mathbf{u}}}_{{F_1}}}, \cdots ,{{\tilde {\mathbf{u}}}_{{F_H}}} \}$, ${\tilde {\mathbf{U}}}_N=\{ {{\tilde {\mathbf{u}}}_{{N_1}}}, \cdots ,{{\tilde {\mathbf{u}}}_{{N_H}}} \}$, ${\tilde {\mathbf{U}}}_A=\{ {{\tilde {\mathbf{u}}}_{{A_1}}}, \cdots ,{{\tilde {\mathbf{u}}}_{{A_H}}} \}$ and $L$ is the stance label set.

Finally, we combine $score_{{s_i}}^{sem}$ and $score_{{s_i}}^{dis}$ to guide selecting the index of the maximum as the predicted label:
\begin{equation}
    scor{e_{{s_i}}} = score_{{s_i}}^{sem} + score_{{s_i}}^{dis}
\end{equation}

\section{Experiments}
In this section, we perform extensive experiments on two datasets for stance detection to assess the effectiveness of our proposed method CoSD. We begin by introducing the two datasets in Section \ref{datasets}. Then we provide details regarding the implementation and baseline methods in Section \ref{details} and Section \ref{baseline}, respectively. Finally, the overall results are presented in Section \ref{results}.

\subsection{Datasets} \label{datasets}
We evaluate our method on SemEval-2016 Task 6 Sub-task A dataset \cite{mohammad-etal-2016-semeval} and UKP dataset \cite{DBLP:conf/emnlp/StabMSRG18}.

The SemEval-2016 dataset consists of 4,163 English tweets on 5 different targets, including “Atheism (AT)”, “Climate Change is a real Concern (CC)”, “Feminist Movement (FM)”, “Hillary Clinton (HC)” and “Legalization of Abortion (LA)”. The dataset is annotated to detect whether a tweet is Favor, None, or Against to a given target. Like previous work, we adopt the official train/test set split. Because the task didn't provide the official validation set, we split the train set into train and validation sets in a ratio of 5:1. The statistics of the SemEval-2016 dataset are shown in Table \ref{tab:1}.

The UKP dataset consists of 25,492 argument sentences. Each sentence is assigned a topic and an annotated label (Argument\_for, NoArgument, Argument\_against) to the topic. There are 8 targets in the dataset: “Abortion (AB)”, “Cloning (CL)”, “Death Penalty (DP)”, “Gun Control (GC)”, “Marijuana Legalization (ML)”, “Minimum Wage (MW)”, “Nuclear Energy (NE)” and “School Uniforms (SU)”. We adopt the train, validation, and test splits provided by the original paper. The statistics of the UKP dataset are shown in Table \ref{tab:2}.

\begin{table}[!t]
\begin{center}
    \caption{Statistics of the SemEval-2016 dataset} 
    \label{tab:1}
    \resizebox{\columnwidth}{!}{
    \begin{tabular}{c|cccc|cccc}
    \toprule
    \textbf{Target} & \textbf{Train} & \textbf{Favor} & \textbf{None} & \textbf{Against} & \textbf{Test} & \textbf{Favor} & \textbf{None} & \textbf{Against} \\
    \midrule
    \textbf{AT} & 513 & 92 & 117 & 304 & 220 & 32 & 28 & 160 \\
    \textbf{CC} & 395 & 212 & 168 & 15 & 169 & 123 & 35 & 11 \\
    \textbf{FM} & 664 & 210 & 126 & 328 & 285 & 58 & 44 & 183 \\
    \textbf{HC} & 689 & 118 & 178 & 393 & 295 & 45 & 78 & 172 \\
    \textbf{LA} & 653 & 121 & 177 & 355 & 280 & 46 & 45 & 189 \\
    \midrule
    \textbf{Total} & 2914 & 753 & 766 & 1395 & 1249 & 304 & 230 & 715 \\
   \bottomrule
    \end{tabular}}
\end{center}
\vspace{-3mm}
\end{table}

\begin{table}[!t]
\begin{center}
    \caption{Statistics of the UKP dataset}
    \label{tab:2}
    \resizebox{\columnwidth}{!}{
    \begin{tabular}{c|cccc|cccc|cccc}
    \toprule
    \textbf{Target} & \textbf{Train} & \textbf{For} & \textbf{No} & \textbf{Against} & \textbf{Val} & \textbf{For} & \textbf{No} & \textbf{Against} & \textbf{Test} & \textbf{For} & \textbf{No} & \textbf{Against} \\
    \midrule
    \textbf{AB} & 2827 & 490 & 1746 & 591 & 315 & 54 & 195 & 66 & 787 & 136 & 486 & 165 \\
    \textbf{CL} & 2187 & 508 & 1075 & 604 & 243 & 56 & 120 & 67 & 609 & 142 & 299 & 168 \\
    \textbf{DP} & 2627 & 316 & 1522 & 789 & 293 & 38 & 165 & 90 & 731 & 103 & 396 & 232 \\
    \textbf{GC} & 2404 & 566 & 1359 & 479 & 268 & 63 & 152 & 53 & 669 & 158 & 378 & 133 \\
    \textbf{ML} & 1780 & 422 & 908 & 450 & 198 & 47 & 101 & 50 & 497 & 118 & 253 & 126 \\
    \textbf{MW} & 1778 & 414 & 968 & 396 & 198 & 46 & 108 & 44 & 497 & 116 & 270 & 111 \\
    \textbf{NE} & 2573 & 436 & 1524 & 613 & 286 & 48 & 170 & 68 & 717 & 122 & 424 & 171 \\
    \textbf{SU} & 2165 & 392 & 1248 & 525 & 241 & 44 & 139 & 58 & 602 & 109 & 347 & 146 \\
    \midrule
    \textbf{Total} & 18341 & 3544 & 10350 & 4447 & 2042 & 396 & 1150 & 496 & 5109 & 1004 & 2853 & 1252 \\
   \bottomrule
    \end{tabular}}
\end{center}
\vspace{-3mm}
\end{table}

\subsection{Implementation Details} \label{details}
In all experiments, we initialize the embedding representation and train using a pre-trained BERT base model (BERT-base-uncased). For collaborative training, Xavier initializer is utilized to initialize the model parameters, except for the embedding representation parameters initialized by BERT as described in Equation \ref{con:equation5}. The embedding dimension is set to 768, while the dimension of the representations $\mathbf{e}^l$ (except $\mathbf{e}^0$) that aggregate the collaboration information of $l$-hop nodes is set to 64. Our method is implemented in PyTorch, and we report the average result of three random trials. The Adam optimizer is employed with a learning rate of 1e-5 for the CPA Module and a learning rate of 0.0001 for BERT. To alleviate overfitting, we apply dropout at a rate of 0.1 to nodes and edges in the heterogeneous topic graph $G$. The batch size is set to 32. For the SemEval-2016 dataset, we set the hop number $l$ to 3 and the topic number $H$ to 5, while 2 hops and 5 topics are set in UKP.

Following previous stance detection works, we use official evaluation metrics of SemEval16, which is the macro-average of F1 score (denoted as $MacF_{avg}$) and the micro-average of F1 score (denoted as $MicF_{avg}$) to evaluate the performance of our method on stance detection task. Since we are more interested in the Favor class and the Against class, the formula for calculating $F_{avg}$ is defined as:
\begin{equation}
\begin{aligned}
    {F_{avg}} &= \frac{{{F_{favor}} + {F_{against}}}}{2} \\
    {F_{favor}} &= \frac{{2 \times {P_{favor}} \times {R_{favor}}}}{{{P_{favor}} + {R_{favor}}}} \\
    {F_{against}} &= \frac{{2 \times {P_{against}} \times {R_{against}}}}{{{P_{against}} + {R_{against}}}}
\end{aligned}
\end{equation}
where $P$ and $R$ denote precision and recall respectively. $MacF_{avg}$: first get the ${F_{avg}}$ for each target and then take the average of all the ${F_{avg}}$. $MicF_{avg}$: first calculate ${F_{favor}}$ and ${F_{against}}$ across all the targets and then take their average.


\begin{table*}[!t]
\begin{center}
\caption{Stance detection results on the UKP dataset}
\label{tab:4}
\resizebox{0.8\linewidth}{!}{
\begin{tabular}{l|cccccccc|cc}
\toprule
\textbf{Method} & AB & CL & DP & GC & ML & MW & NE & SU & $MacF_{avg}$ & $MicF_{avg}$ \\
\midrule
${BER{T}_{SEP}}$
& 0.4909 & 0.6690 & 0.5242 & 0.5162 & 0.6631 & 0.6491 & 0.5854 & 0.6342 & 0.5915 & 0.5973 \\
${BER{T}_{MEAN}}$
& 0.5376 & 0.6855 & 0.5416 & 0.5112 & 0.6518 & 0.6640 & 0.5776 & 0.6327 & 0.6003 & 0.6019 \\
${BER{T}_{TAN}}$
& 0.4057 & 0.6936 & 0.5163 & 0.5098 & 0.6347 & 0.6410 & 0.5670 & 0.6408 & 0.5761 & 0.5794 \\
${{\rm CKD}_{BERT}}$
& 0.5019 & 0.7182 & 0.5665 & 0.4912 & 0.6276 & 0.6782 & 0.5823 & 0.6725 & 0.6094 & 0.6048 \\
${{\rm CKD}_{BERTweet}}$
& 0.5779 & 0.7182 & \underline{0.6198} & 0.5564 & 0.6937 & \underline{0.7182} & 0.6653 & 0.7245 & 0.6577 & 0.6593 \\
\midrule
Stancy
& 0.5176 & 0.6752 & 0.5671 & 0.5024 & 0.6535 & 0.6615 & 0.5832 & 0.6291 & 0.5987 & 0.6006 \\
TAPD
& 0.5487 & 0.7186 & 0.5716 & 0.5173 & 0.6701 & 0.6995 & 0.5903 & 0.6410 & 0.6196 & 0.6215 \\
\midrule
${{\rm TSE}_{Bi-LSTM}}$
& 0.4379  & 0.5563 & 0.3736 & 0.4350 & 0.4661 & 0.4812 & 0.4617 & 0.4919 & 0.4630 & 0.4757 \\
${{\rm TSE}_{Bi-LSTM+Multi-task}}$
& 0.4008 & 0.5682 & 0.4206 & 0.4585 & 0.5270 & 0.5120 & 0.4912 & 0.4630 & 0.4901 & 0.4801 \\
${{\rm TSE}_{BERTweet}}$
& 0.5911 & \underline{0.7255} & 0.5848 & 0.5637 & \underline{0.7066} & 0.6780 & \underline{0.6861} & \textbf{0.7491} & \underline{0.6606} & 0.6583 \\
${{\rm TSE}_{BERTweet+Multi-task}}$
& \underline{0.6058} & 0.7071 & 0.6169 & \underline{0.5876} & 0.7062 & 0.6601 & 0.6579 & \underline{0.7358} & 0.6597 & \underline{0.6600} \\
\midrule
\textbf{CoSD} & \textbf{0.6782} & \textbf{0.7425} & \textbf{0.6832} & \textbf{0.6138} & \textbf{0.7270} & \textbf{0.7493} & \textbf{0.7183} & 0.6343 & \textbf{0.6933} & \textbf{0.6940} \\
${\Delta_{SOTA}}$ & \color{red}↑7.24\% & \color{red}↑1.7\% & \color{red}↑6.34\% & \color{red}↑2.62\% & \color{red}↑2.04\% & \color{red}↑3.11\% & \color{red}↑3.22\% & - & \color{red}↑3.27\% & \color{red}↑3.4\% \\
\bottomrule
\end{tabular}}
\vspace{-3mm}
\end{center}
\end{table*}

\subsection{Baselines} \label{baseline}
Our approach diverges from conventional classification task frameworks, which primarily emphasize capturing collaborative signals between texts to identify similar ones and provide classification support for the target text. Consequently, we have selected several established classification model frameworks as baselines. These methods primarily focus on stance detection from two perspectives: one being model-based and the other relying on supplementary information. Building upon this foundation, we have chosen 16 representative and state-of-the-art baselines in stance detection for comparison to showcase the effectiveness of our CoSD. The selected baselines are enumerated as follows:

\noindent\textbf{Model-based:} mainly leveraging the relationship between texts and their respective targets.
\begin{itemize}
\item \textbf{TAN} \cite{Du2017StanceCW}: An LSTM-attention-based model that leverages the attention module to incorporate target-specific information into stance detection.
\item \textbf{PNEM} \cite{DBLP:conf/naacl/SiddiquaCA19}: An ensemble model that adopted two LSTM-attention-based models to learn long-term dependencies and a multi-kernel convolution to extract the higher-level tweet representation.
\item \textbf{$\mathbf{BER}{\mathbf{T}_{SEP}}$} \cite{Devlin2019BERTPO}: A pre-trained language model that predicted the stance by appending a linear classification layer to the hidden representation of [CLS] token. Fine-tune the BERT on the stance detection.
\item \textbf{$\mathbf{BER}{\mathbf{T}_{MEAN}}$} \cite{Ma2019UniversalTR}: A pre-trained language model that predicted the stance by appending a linear classification layer to the mean hidden representation over all tokens. Fine-tune the BERT on the stance detection.
\item \textbf{$\mathbf{BER}{\mathbf{T}_{TAN}}$} \cite{DBLP:reference/ml/ZhengW10b}: A variant of TAN, which replaces the LSTM features with BERT generated word embeddings.
\item \textbf{S-MDMT} \cite{Wang2021SolvingSD}: A BERT-based model that applies adversarial learning to capture stance-related features shared by all targets and combines target descriptors to learn stance-informative features associated with specific targets.
\item \textbf{CKD} \cite{li-caragea-2023-distilling}: A knowledge distillation method for stance detection in which a student model is taken as a new teacher model to transfer knowledge to a new fresh student model over multiple generations. 
\end{itemize}

\noindent\textbf{Supplementary information-based:} including external knowledge enhancement and joint task learning.

\subsubsection{\textbf{external knowledge enhancement}}
\begin{itemize}
\item \textbf{Stancy} \cite{popat-etal-2019-stancy}: A BERT-based model that leverages BERT representations trained over massive external corpora and utilizes consistency constraint to model target and text.
\item \textbf{CKEMN} \cite{9179107}: A commonsense knowledge enhanced memory network for stance detection.
\item \textbf{RelNet} \cite{DBLP:conf/ccks/ZhangYZ021}: A multiple knowledge enhanced framework for stance detection using BERT.
\item \textbf{TAPD} \cite{10.1145/3477495.3531979}: A prompt-based fine-tuning method for stance detection, which designs target-aware prompts and distills PLMs with multiple prompts.
\end{itemize}

\subsubsection{\textbf{joint task learning}}
\begin{itemize}
\item \textbf{JOINT} \cite{DBLP:journals/fcsc/SunWLZZ19}: A joint model that exploited sentiment information to improve stance detection.
\item \textbf{AT-JSS-Lex} \cite{DBLP:conf/emnlp/LiC19a}: A multi-task framework that used a sentiment lexicon and constructed a stance lexicon to guide target-specific attention mechanism. Besides, it took sentiment classification as an auxiliary task.
\item \textbf{MTIN} \cite{chai-etal-2022-improving}: A novel multi-task interaction network for improving the performance of stance detection and sentiment analysis tasks simultaneously.
\item \textbf{TSE} \cite{li-etal-2023-new}: A two-stage framework that first identifies the relevant target in the text and then detects the stance given the predicted target and text.
\end{itemize}

\noindent\textbf{Similarity-based:} We also compare CoSD with a state-of-the-art model considering the correlation between text and target.
\begin{itemize}
\item \textbf{KNN-TACL} \cite{10.1145/3583780.3615085}: A topic-aware model that designs a pretext task to mine the topic associations and model the topic association for contrastive learning.
\end{itemize}

\begin{table}[!t]
\begin{center}
\caption{Stance detection results on SemEval-2016}
\label{tab:3}
\resizebox{\columnwidth}{!}{
\begin{tabular}{l|ccccc|cc}
\toprule
\textbf{Method} & AT & CC & FM & HC & LA & $MacF_{avg}$ & $MicF_{avg}$ \\
\midrule
TAN
& 0.5933 & 0.5359 & 0.5577 & 0.6538 & 0.6372 & 0.5956 & 0.6879 \\
PNEM
& 0.6773 & 0.4427 & 0.6676 & 0.6028 & 0.6423 & 0.6065 & 0.7211 \\
${BER{T}_{SEP}}$
& 0.6867 & 0.4414 & 0.6166 & 0.6234 & 0.5860 & 0.5909 & 0.6951 \\
${BER{T}_{MEAN}}$
& 0.6944 & 0.5247 & 0.5922 & 0.6461 & 0.6630 & 0.6241 & 0.7092 \\
${BER{T}_{TAN}}$
& 0.6551 & 0.5855 & 0.5830 & 0.6431 & 0.6358 & 0.6205 & 0.7028 \\
S-MDMT
& 0.6950 & 0.5249 & 0.6378 & 0.6720 & 0.6719 & 0.6403 & 0.7270 \\
${{\rm CKD}_{BERT}}$
& \underline{0.7532} & 0.4502 & 0.5238 & 0.6162 & 0.6456 & 0.5978 & 0.7082 \\
${{\rm CKD}_{BERTweet}}$
& 0.7088 & 0.4556 & 0.6190 & 0.7194 & 0.6731 & 0.6352 & 0.7440 \\
\midrule
Stancy
& 0.6985 & 0.5347 & 0.6167 & 0.6470 & 0.6342 & 0.6262 & 0.7177 \\
CKEMN
& 0.6269 & 0.5352 & 0.6125 & 0.6419 & 0.6419 & 0.6117 & 0.6974 \\
RelNet
& 0.7055 & 0.5720 & 0.6125 & 0.6233 & 0.6365 & 0.6306 & 0.7106 \\
TAPD
& 0.7387 & \underline{0.5932} & 0.6393 & 0.7001 & 0.6723 & 0.6687 & 0.7480 \\
\midrule
JOINT
& 0.6678 & 0.506 & 0.5935 & 0.6247 & 0.6158 & 0.6016 & 0.6922 \\
AT-JSS-Lex
& 0.6922 & 0.5918 & 0.6149 & 0.6833 & 0.6841 & 0.6533 & 0.7233 \\
MTIN
& - & - & - & - & - & 0.649 & 0.703 \\
${{\rm TSE}_{Bi-LSTM}}$
& 0.5323 & - & 0.5077 & 0.5338 & 0.5111 & 0.5212 & 0.5275 \\
${{\rm TSE}_{Bi-LSTM+Multi-task}}$
& 0.5388  & - & 0.5382 & 0.5355 & 0.6690 & 0.5704 & 0.5744 \\
${{\rm TSE}_{BERTweet}}$
& 0.6787 & - & 0.6592 & \underline{0.7607} & 0.7207 & 0.7048 & 0.7032 \\
${{\rm TSE}_{BERTweet+Multi-task}}$
& 0.7090 & - & \textbf{0.6903} & 0.7601 & \underline{0.7400} & \underline{0.7248} & 0.7226 \\
\midrule
KNN-TACL
& 0.7433 & 0.4538 & 0.6540 & 0.7102 & 0.6741 & 0.6471 & \underline{0.7490} \\
\midrule
\textbf{CoSD} & \textbf{0.8102} & \textbf{0.6833} & \underline{0.6896} & \textbf{0.7635} & \textbf{0.7729} & \textbf{0.7439} & \textbf{0.8032} \\
${\Delta_{SOTA}}$ & \color{red}↑5.7\% & \color{red}↑9.01\% & - & \color{red}↑0.28\% & \color{red}↑3.29\% & \color{red}↑1.91\% & \color{red}↑5.52\% \\
\bottomrule
\end{tabular}}
\vspace{-3mm}
\end{center}
\end{table}

The results for ${BER{T}_{SEP}}$, ${BER{T}_{MEAN}}$, ${BER{T}_{TAN}}$ and Stancy are from \cite{10.1145/3477495.3531979}, the results for CKD and TSE are obtained through our experiments, while the results for other models are directly extracted from their respective original papers. \textbf{${\mathbf{CKD}_{BERT}}$} and \textbf{${\mathbf{CKD}_{BERTweet}}$} refer to the pretrained BERT and BERTweet models employed as teacher and student models for CKD, respectively. For TSE, \textbf{${\mathbf{TSE}_{Bi-LSTM}}$} and \textbf{${\mathbf{TSE}_{BERTweet}}$} indicate  the use of Bi-LSTM and BERTweet\cite{DBLP:conf/acl/LiC21} as baselines in stage one, respectively. BERTweet\cite{DBLP:conf/emnlp/NguyenVN20} is a pre-trained language model trained following the training procedure of RoBERTa\cite{Liu2019RoBERTaAR}. ${Multi-task}$ indicates whether target prediction is utilized as an auxiliary task in the second stage. Note that according to the Settings in the original paper, TSE removes target "CC" in the SemEval dataset.

\subsection{Overall Results} \label{results}
Table \ref{tab:4} and Table \ref{tab:3} present experimental results on two datasets, highlighting the best performances in bold and the second-best performing methods underlined. ${\Delta_{SOTA}}$ denotes the comparison of our method with the current state-of-the-art results. It is evident that our approach performs effectively on both datasets, surpassing baseline models in overall metrics $MacF_{avg}$ and $MicF_{avg}$. Specifically, on the SemEval-2016 dataset, our method achieves improvements of 1.91\% and 5.52\% in $MacF_{avg}$ and $MicF_{avg}$, respectively. Similarly, on the UKP dataset, it demonstrates improvements of 3.27\% and 3.4\% over $MacF_{avg}$ and $MicF_{avg}$. Furthermore, our method achieves either the best or comparable results for each target.

Model-based methods, which emphasize the relationship between texts and targets, significantly enhance the accurate comprehension of text semantics. In CKD, knowledge distillation iteratively synthesizes the knowledge gleaned from the base model, resulting in a substantial enhancement in text understanding.

External knowledge enhancement methods, whether utilizing vast external corpora (Stancy), commonsense knowledge (CKEMN), or extracting rich knowledge from pre-trained language models (TAPD) through prompts, indeed offer the potential to augment texts or targets with additional background information, thereby enhancing performance. For instance, in the case of the highly imbalanced dataset with the target "CC" in SemEval-2016, which is addressed by TSE through removal, prompt learning supplements it with abundant missing information, allowing TAPD to perform effectively. However, our method significantly enhances results by leveraging collaborative signals to identify similar texts and extract correlations between targets, without relying on external knowledge. Compared with traditional classification models that are agnostic to structure and insensitive to inter-stance
correlation, this underscores the significance of extracting collaborative signals in improving performance.

Joint task learning methods can offer supplementary information from different perspectives and levels to texts or targets, thereby exerting varying degrees of positive impact on overall performance or specific targets. In TSE, the target extracted in the first stage plays a pivotal role in guiding stance detection results. When comparing ${{\rm TSE}_{BERTweet}}$ and ${{\rm TSE}_{BERTweet+Multi-task}}$, the inclusion of the ${Multi-task}$ setting exhibits minimal difference in overall effectiveness. However, it still contributes to improving the performance of datasets lacking information and requiring supplementation with external knowledge.

In the case of CKD and TSE, the exceptional performance on both datasets underscores the superior capability of Pre-trained Language Models (PLMs) in extracting semantic information. Notably, the overall performance metric $MicF_{avg}$ of KNN-TACL, which leverages textual correlation on SemEval-2016, highlights the significance of considering similar samples. This underscores the importance of utilizing collaborative information between texts in our approach to alleviate the model's structure insensitivity.

\section{Analysis and Case Study}
In this section, we conduct a comprehensive analysis of the design of our proposed method. Initially, we examine the influence of various components of our method as described in Section \ref{ablation}. Subsequently, we evaluate the method's performance under different hyperparameter configurations outlined in Section \ref{hyperparameter}, followed by a visual classification analysis in Section \ref{visualization}. Finally, we present case studies and visualizations in Sections \ref{case} and \ref{case visualization}, respectively.

\begin{table}[!t]
\begin{center}
\caption{Results of ablation study on SemEval} 
\label{tab:5}
\resizebox{\columnwidth}{!}{
\begin{tabular}{l|ccccc|cc}
\toprule
\textbf{Method} & AT & CC & FM & HC & LA & $MacF_{avg}$ & $MicF_{avg}$ \\
\midrule
CoSD $\neg$ CT-SVM & 0.4211 & 0.4212 & 0.3919 & 0.3683 & 0.4038 & 0.4013 & 0.6527 \\
CoSD $\neg$ BERT & 0.7545 & 0.6509 & 0.6842 & 0.6237 & 0.65   & 0.6727 & 0.6898 \\
CoSD $\neg$ LDA & 0.686  & 0.4291 & 0.5846 & 0.6464 & 0.6323 & 0.5957 & 0.6949 \\
\textbf{CoSD} & \textbf{0.8102} & \textbf{0.6833} & \textbf{0.6896} & \textbf{0.7635} & \textbf{0.7729} & \textbf{0.7439} & \textbf{0.8032} \\
\bottomrule
\end{tabular}}
\end{center}
\vspace{-3mm}
\end{table}

\begin{table}[!t]
\begin{center}
\caption{Results of ablation study on UKP} 
\label{tab:6}
\resizebox{\columnwidth}{!}{
\begin{tabular}{l|cccccccc|cc}
\toprule
\textbf{Method} & AB & CL & DP & GC & ML & MW & NE & SU & $MacF_{avg}$ & $MicF_{avg}$ \\
\midrule
CoSD $\neg$ CT-SVM & 0.3818 & 0.3184 & 0.3514 & 0.361  & 0.3373 & 0.352  & 0.3716 & 0.4735 & 0.3684 & 0.3766 \\
CoSD $\neg$ BERT & 0.5619 & 0.6295 & 0.5546 & 0.53   & \textbf{0.7302} & 0.6561 & 0.5447 & 0.6076 & 0.6018 & 0.5958 \\
CoSD $\neg$ LDA & \textbf{0.6918} & 0.7378 & 0.6698 & 0.5998 & 0.7296 & 0.7397 & 0.6996 & 0.6287 & 0.6911 & 0.6868 \\
\textbf{CoSD} & 0.6782 & \textbf{0.7425} & \textbf{0.6832} & \textbf{0.6138} & 0.727  & \textbf{0.7493} & \textbf{0.7183} & \textbf{0.6343} & \textbf{0.6933} & \textbf{0.694} \\
\bottomrule
\end{tabular}}
\end{center}
\vspace{-3mm}
\end{table}

\subsection{Ablation Study} \label{ablation}
To validate the effectiveness of the different modules of CoSD, we compare them with the following variants:

\textbf{CoSD $\neg$ CT-SVM} eliminates the contrastive graph collaborative training process and substitutes it with SVM. In this variant, the implicit topic distribution generated by LDA is directly input into SVM for classification. This model fails to capture collaborative signals.

\textbf{CoSD $\neg$ BERT} removes BERT from the inference stage, resulting in a loss of semantic perception in the model.

\textbf{CoSD $\neg$ LDA} removes LDA from the inference stage, leading to a lack of distributed perception.

Table \ref{tab:5} and Table \ref{tab:6} show the results on SemEval and UKP datasets, respectively. CoSD $\neg$ CT-SVM's poor performance on the two datasets indicates that the collaborative Contrastive Graph Training plays a significant role in the whole model, and the CPA module well obtains the collaboration information between the implicit topic and the text for classification.

When comparing CoSD $\neg$ BERT and CoSD $\neg$ LDA to CoSD, there is a different magnitude of decline in both datasets, which shows the ability and importance of BERT in acquiring text semantic information, and the guiding role of collaborative signals in text classification. Compared with CoSD $\neg$ BERT, CoSD $\neg$ LDA has a larger drop on the SemEval dataset, which indicates that there is richer collaboration information on the SemEval dataset. On the contrary, semantic information plays a relatively important role in UKP.

For the UKP dataset, target "AB" and target "ML" achieve best results respectively on CoSD $\neg$ LDA and CoSD $\neg$ BERT, which indicates that these two targets have a strong preference for semantic information and collaboration signals, while CoSD pays equal attention to both parts of information, which leads to its effect being somewhat decreased instead.

\subsection{Hyperparameter Sensitivity} \label{hyperparameter}
\begin{figure}[!t]
	\centering
    \subfigure[SemEval-2016]{
        \includegraphics[width=0.45\columnwidth]{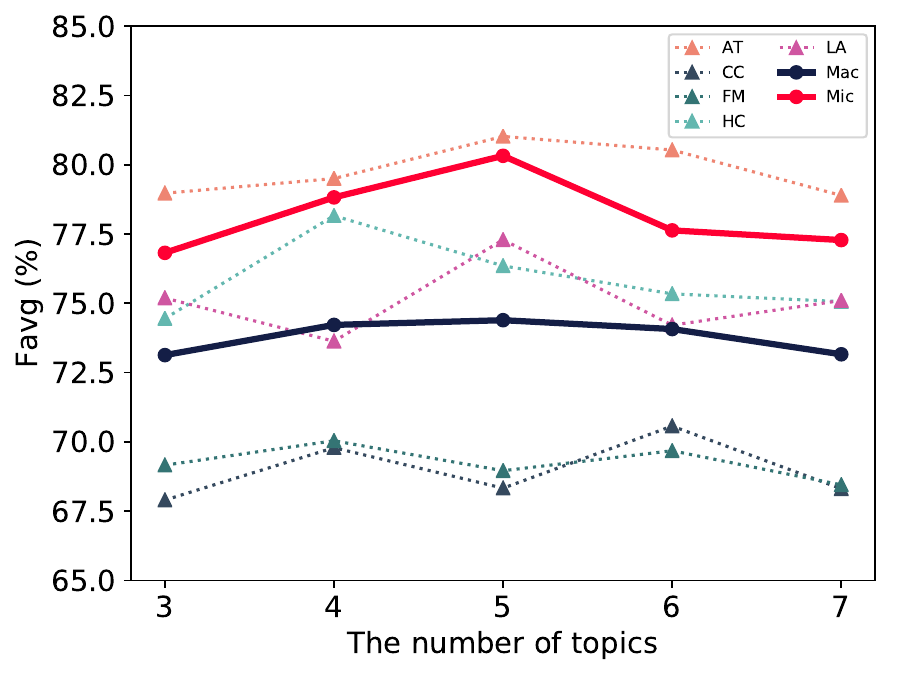}}
	\subfigure[UKP]{
        \includegraphics[width=0.45\columnwidth]{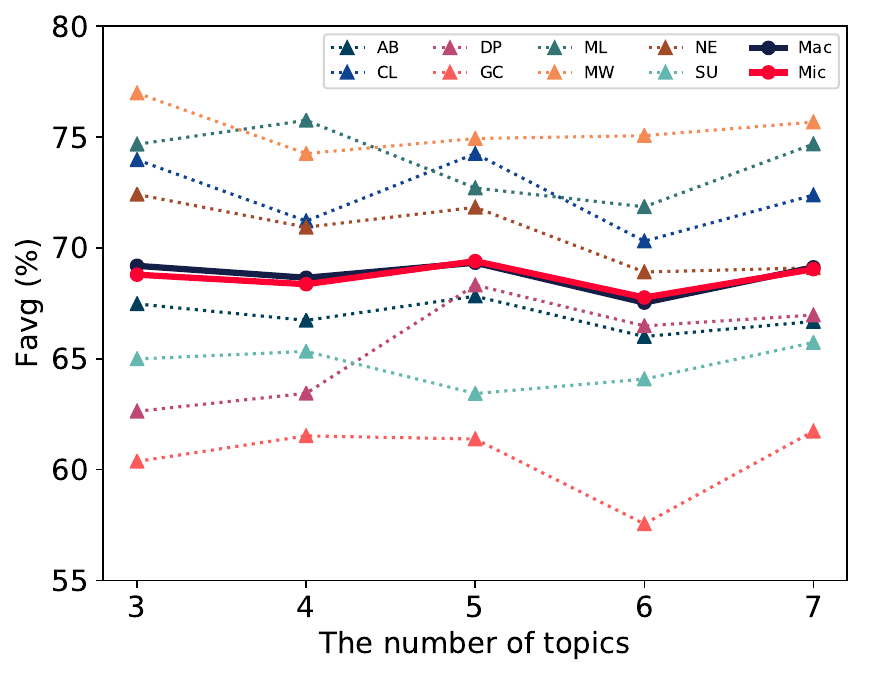}}
        \vspace{-2mm}
	\caption{Performance of CoSD under different values of the parameter topic numbers $H$}
    \label{fig:topics}
\end{figure}

\begin{figure}[!t]
	\centering
    \subfigure[SemEval-2016]{
        \includegraphics[width=0.45\columnwidth]{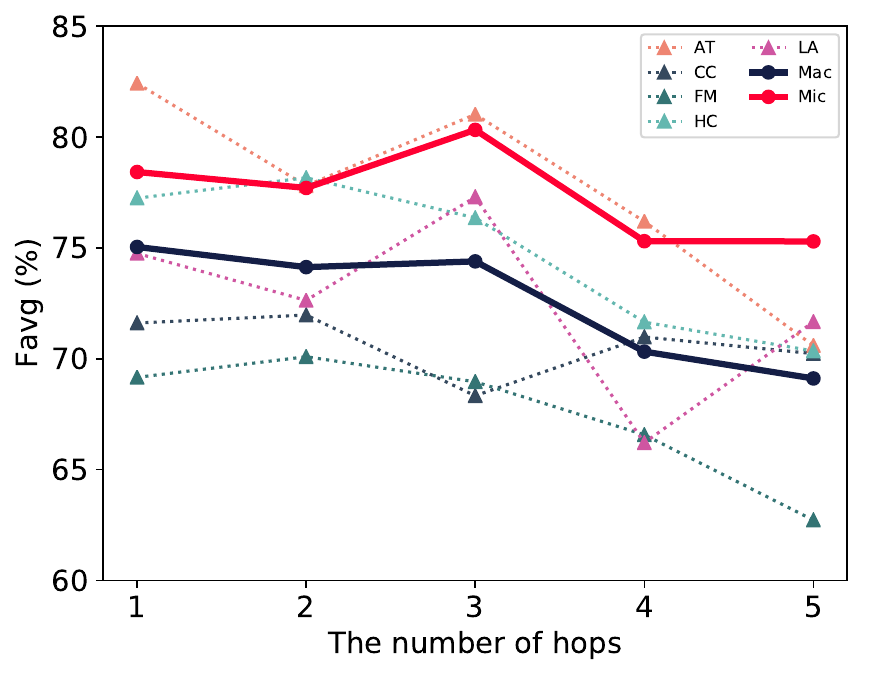}}
	\subfigure[UKP]{
        \includegraphics[width=0.45\columnwidth]{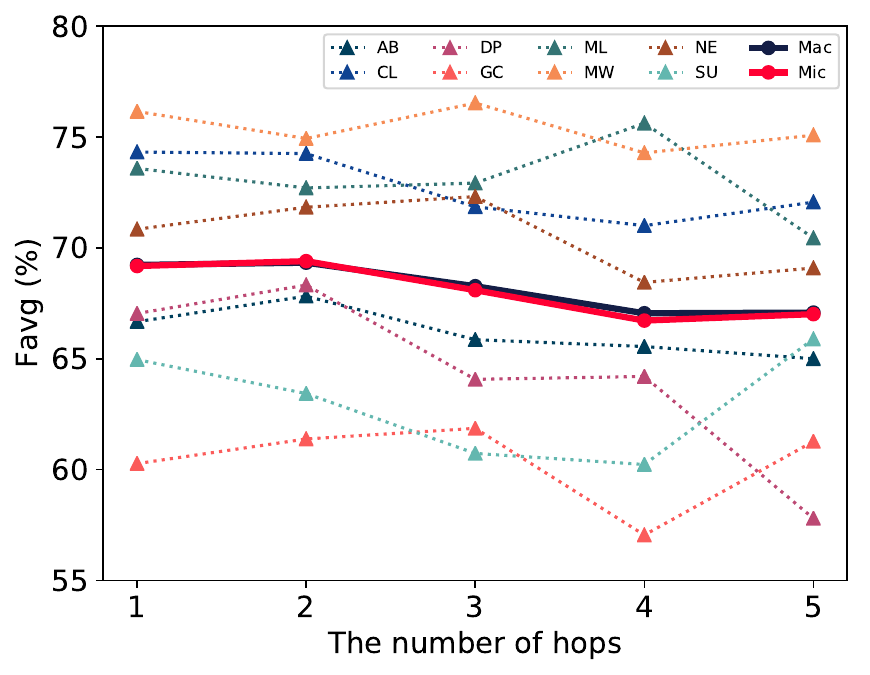}}
        \vspace{-2mm}
	\caption{Performance of CoSD under different values of the parameter hop numbers $l$}
    \label{fig:hops}
\end{figure}

Two parameters, the number of implicit topics ($H$) and the number of hops considered in the CPA module ($l$), are selected for analysis, and a series of experiments are designed accordingly. The results are depicted in Figure \ref{fig:topics} and Figure \ref{fig:hops}. Given that different combinations of parameter values are selected for different targets to achieve optimal performance, we pay more attention to the effect of all targets on the overall data set. The overall evaluation metrics $MacF_{avg}$ and $MicF_{avg}$ are displayed on the graph using solid lines. Since UKP is relatively balanced, the line charts of the two overall indicators exhibit high coincidences.

In determining the number of topics parameter, prior research suggests that setting too many topics will lead to overfitting in topic models. Employing coherence and perplexity as criteria for assessing the reasonableness of the LDA model, we narrowed down the selection range for the number of topics to 3 to 7. The overall effect exhibited an initial increase followed by a decrease. According to Figure \ref{fig:topics}, we ultimately selected 5 topics for both SemEval and UKP datasets to achieve the best overall performance. When setting the number of topics to 5, target “LA” on SemEval and target “DP” on UKP achieve optimal results, which significantly contributed to the overall performance enhancement.

\begin{table*}[!t]
\begin{center}
\caption{Case Study: $Test$ is the test text, $T_1$ and $T_2$ are the top 2 texts with the most similar representations.}
\label{tab:7}
\resizebox{\textwidth}{!}{
\begin{tabular}{c|l|c}
\toprule
\textbf{Target} & \textbf{Case} & \textbf{Stance}  \\
\midrule
AB & \begin{tabular}[c]{@{}l@{}}$Test$: {\color{green}Constitutional rights} are {\color{green}for all Americans}, not just those in states where legislators refuse to allow {\color{green}abortion}. \\$T_1$: Adding claims on the {\color{green}Equal Protection Clause} to the due process basis for {\color{green}abortion rights} can strengthen the case for those {\color{green}rights in constitutional politics} as well as {\color{green}constitutional law}. \\ $T_2$: When {\color{green}abortion} restrictions reflect or enforce {\color{green}traditional sex-role stereotypes}, {\color{green}equality arguments} insist that such restrictions are suspect and may violate the {\color{green}U.S. Constitution}. \end{tabular} & Favor  \\
\midrule
CL & \begin{tabular}[c]{@{}l@{}}$Test$: {\color{green}Reproductive cloning} can {\color{green}provide genetically related children} for {\color{green}people who can not be helped by} other fertility treatments (i.e., {\color{green}who do not} produce eggs or sperm). \\$T_1$: {\color{green}Reproductive technologies} can allow couples {\color{green}who happen to be affected by} the accidents of infertility or {\color{green}genetic} disease to {\color{green}have healthy children}. \\$T_2$: These are a few examples of how {\color{green}cloning} may {\color{green}provide a genetically  related child} to a {\color{green}person otherwise unable to have one}. \end{tabular} & Favor \\
\midrule
DP & \begin{tabular}[c]{@{}l@{}}$Test$: As you may have read in the arguments, the {\color{green}death penalty} help to {\color{green}curtail future murderers}, thus, we can {\color{green}save more lives}. \\$T_1$: If we have the {\color{green}death sentence}, and {\color{green}deter future murderers}, we {\color{green}spared the lives of future victims} - (the prospective murderers gain, too; they are spared punishment because they were deterred). \\$T_2$: And we must {\color{green}execute murderers} as long as it is merely possible that   their execution {\color{green}protects citizens from future murder}. \end{tabular} & Favor \\
\midrule
GC & \begin{tabular}[c]{@{}l@{}}$Test$: Higher levels of gun ownership do not produce a safer society and often lead to a {\color{green}higher numbers} of {\color{green}deaths} due to {\color{green}gun violence}. \\ $T_1$: {\color{green}Gun violence}, in short, is a structural impediment to true equality for {\color{green}high-crime}, {\color{green}low-income} areas. \\$T_2$: {\color{green}Crime} rates, especially {\color{green}violent crime} rates, are {\color{green}higher} in {\color{green}poorer} neighborhoods. \end{tabular} & Favor \\
\midrule
ML & \begin{tabular}[c]{@{}l@{}}$Test$: The normalization, {\color{red}expanded use}, and increased availability of {\color{red}marijuana} in our communities are {\color{red}detrimental to our youth, to public health, and to the safety of our society}. \\$T_1$: {\color{red}Increased drug use}, {\color{red}negative health effects, and negative effects on families} are all the outcomes of legalizing {\color{red}marijuana}. \\$T_2$: {\color{red}Increased drug use} leads to {\color{red}negative health effects}. \end{tabular} & Against \\
\midrule
MW & \begin{tabular}[c]{@{}l@{}}$Test$: {\color{red}Raising the minimum wage} would {\color{red}increase} housing {\color{red}costs}. \\$T_1$: {\color{red}Raise the min wage} and {\color{red}costs increase}. \\$T_2$: An {\color{red}increase} in consumer prices, driven by companies offsetting {\color{red}increased} labor {\color{red}costs}. \end{tabular} & Against \\
\midrule
NE & \begin{tabular}[c]{@{}l@{}}$Test$: Uranium mining can also {\color{red}damage the environment}. \\$T_1$: It can also cause {\color{red}damage to living things} in and around the {\color{red}plants}. \\$T_2$: Nuclear power {\color{red}threatens the environment and people’s health}. \end{tabular} & Against \\
\midrule
SU & \begin{tabular}[c]{@{}l@{}}$Test$: {\color{red}School uniforms} promote {\color{red}conformity over individuality}. \\$T_1$: In terms of personal development, {\color{red}uniforms} promote {\color{red}conformity rather than individuality}. \\$T_2$: Today's {\color{red}school uniforms} seem more a punitive measure meant to deny students their right to freedom of expression and {\color{red}individuality}. \end{tabular} & Against \\
\bottomrule
\end{tabular}}
\vspace{-3mm}
\end{center}
\end{table*}

In selecting the parameter of hop numbers, we opt for 3 hops on SemEval and 2 hops on UKP to achieve the best overall effect. This observation suggests that the collaboration signal on UKP is weaker compared to SemEval, aligning with the results and explanations in \ref{ablation}. As illustrated in Figure \ref{fig:hops}, one key factor driving the selection of 3 hops for SemEval is the significant improvement observed for target "LA" when the hop number is set to 3, indicating that our method effectively captures sufficient collaborative information for target "LA". In Section \ref{ablation}, target "AB" and target "ML" in UKP achieve the best results respectively with CoSD $\neg$ LDA and CoSD $\neg$ BERT, rather than the CoSD. Figure \ref{fig:hops} suggests a possible explanation: the optimal hop numbers were not selected for target "AB" and "ML". For instance, the optimal value is attained when the hop number for target "AB" is 2 and for target "ML" is 4, while the overall setting is 3 hops.

\subsection{Classification Visualization}  \label{visualization}
To verify the effectiveness of our proposed model CoSD, we visualize the experimental results for some targets.

\begin{figure}[!t]
	\centering
    \subfigure[initial]{
        \includegraphics[width=0.45\columnwidth]{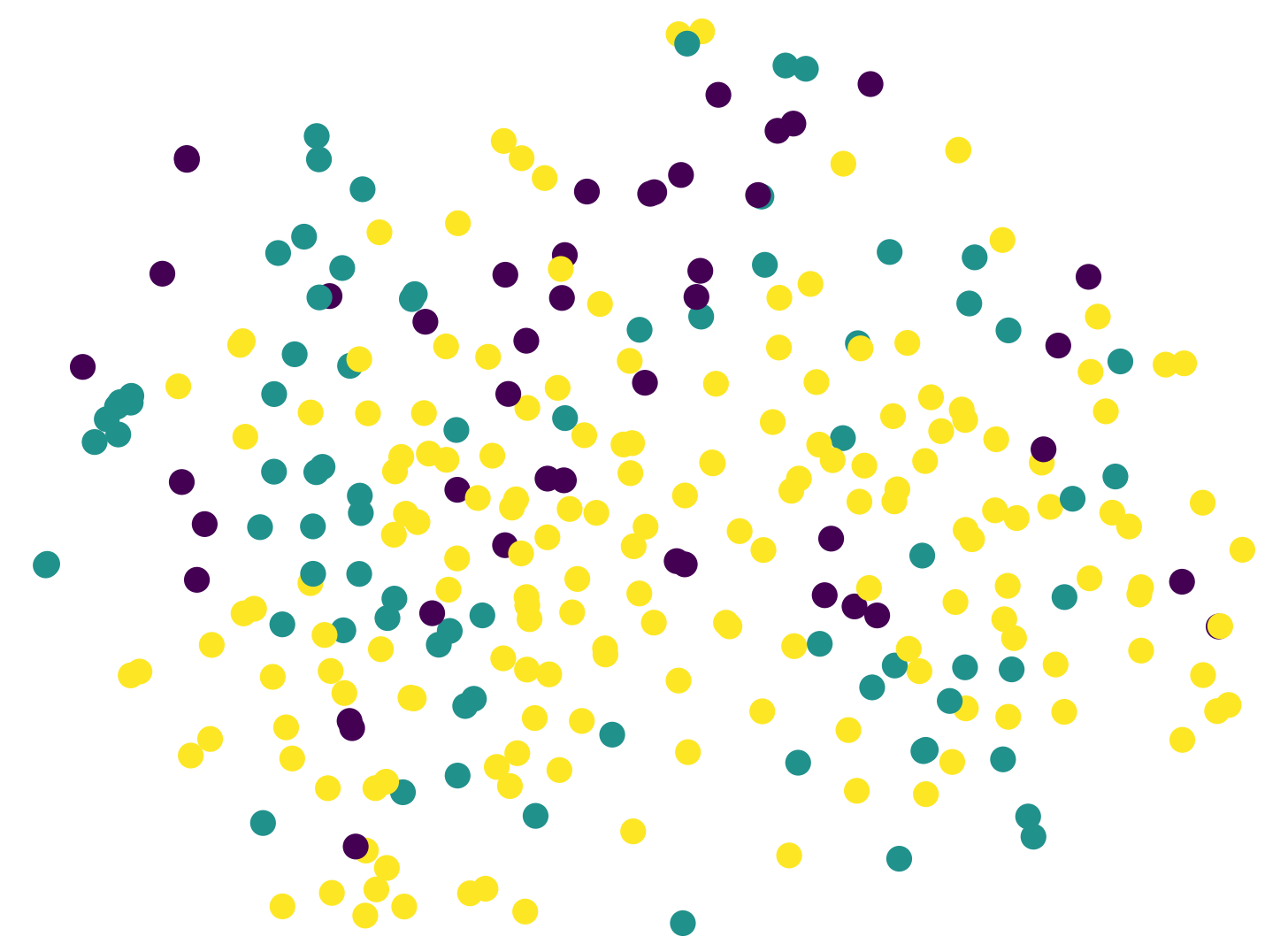}}
    \subfigure[w/o BERT]{
        \includegraphics[width=0.45\columnwidth]{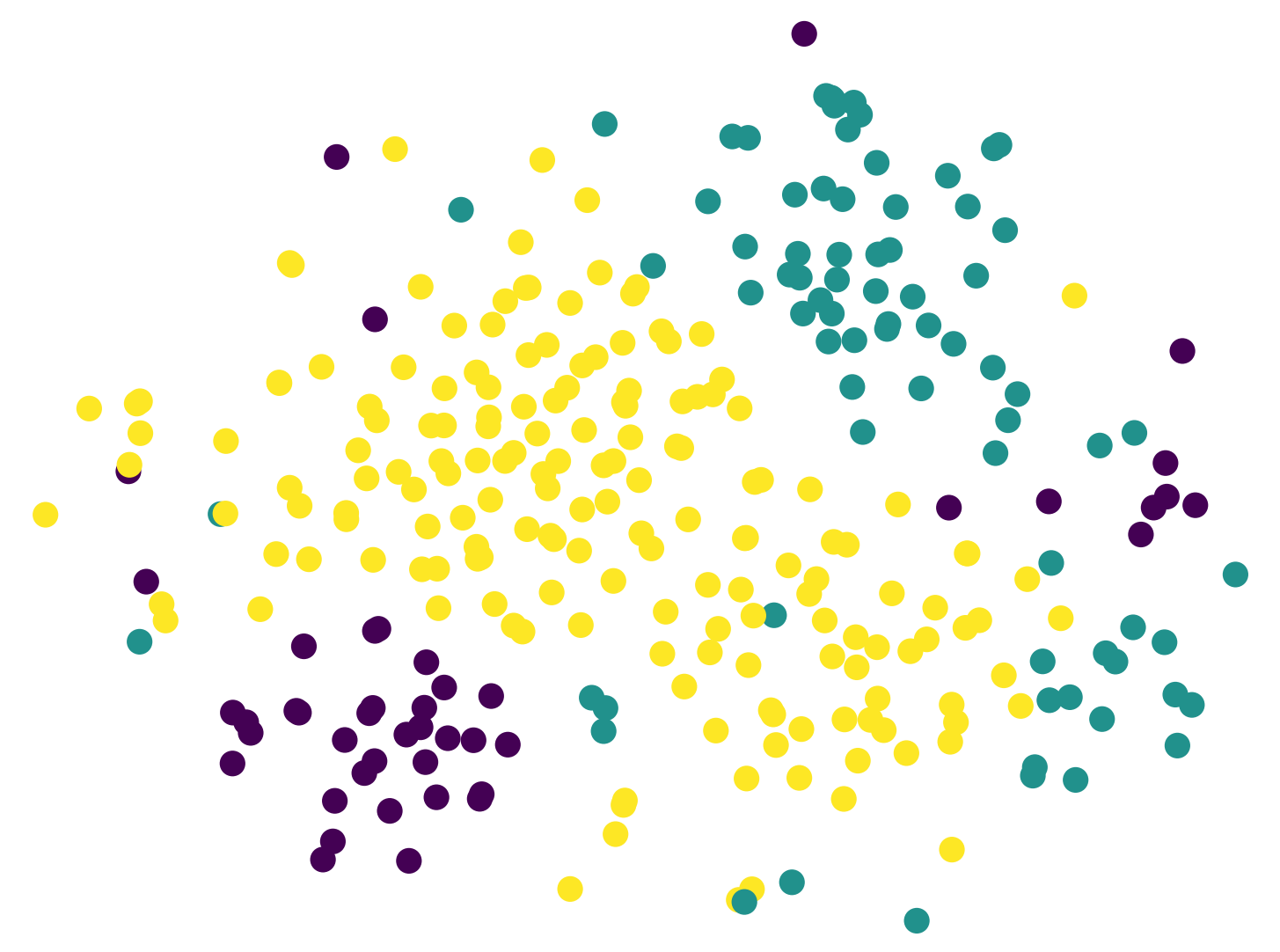}}
    \subfigure[w/o LDA]{
        \includegraphics[width=0.45\columnwidth]{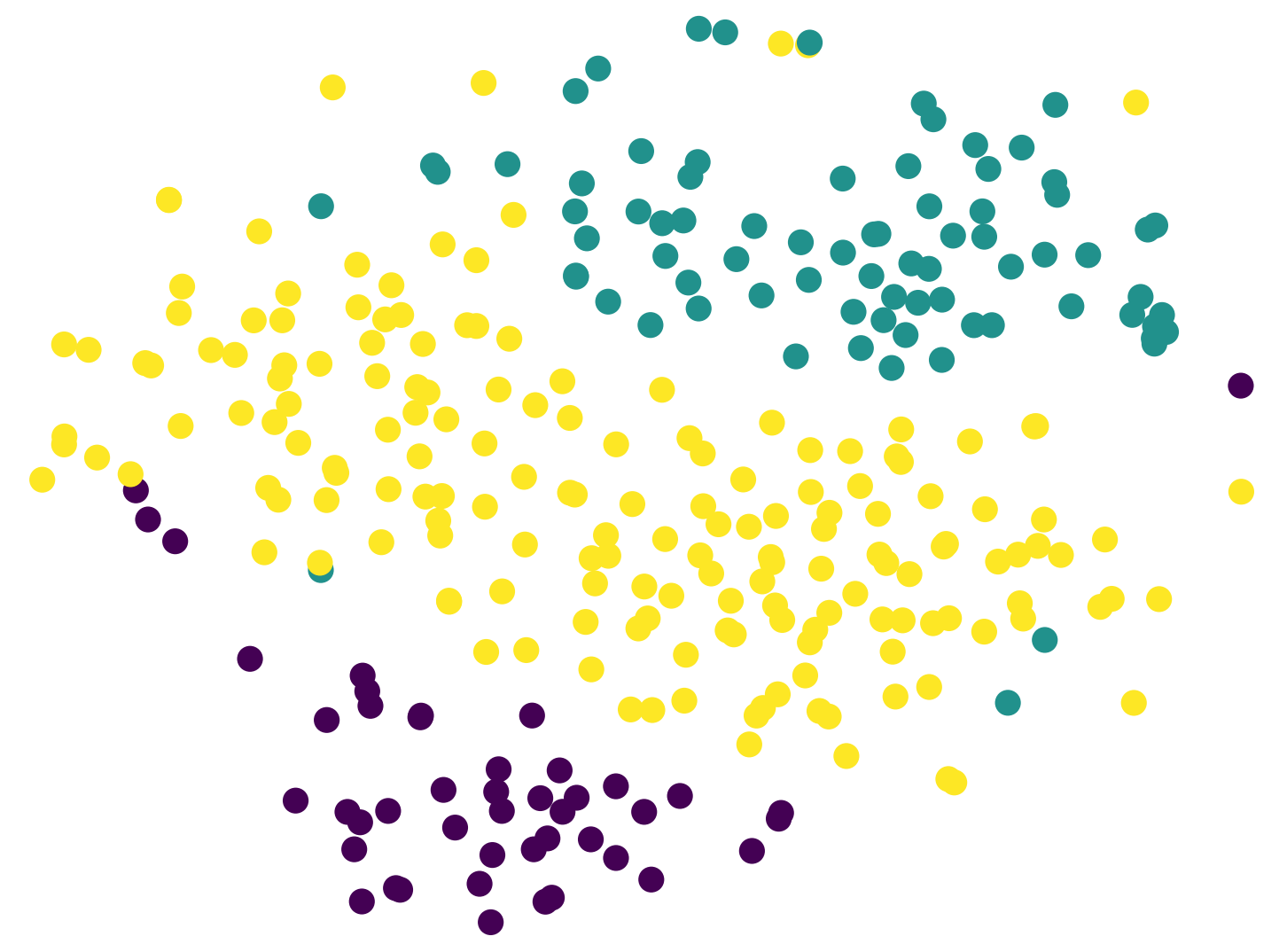}}
    \subfigure[CoSD]{
        \includegraphics[width=0.45\columnwidth]{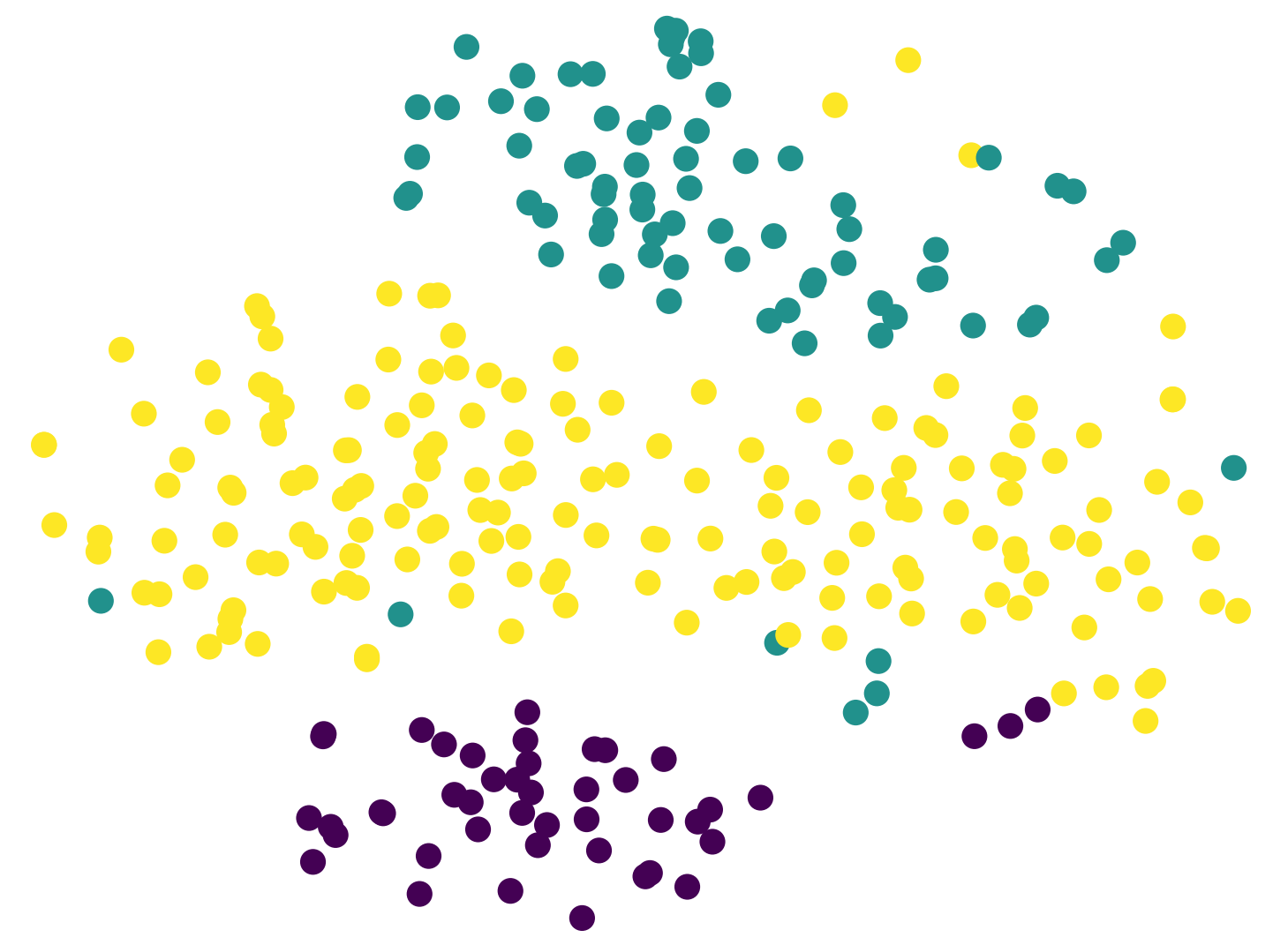}}
        \vspace{-2mm}
	\caption{T-SNE visualization of Target 'HC' on SemEval-2016}
    \label{fig:t-sne1}
\end{figure}

\begin{figure}[!t]
	\centering
    \subfigure[initial]{
        \includegraphics[width=0.45\columnwidth]{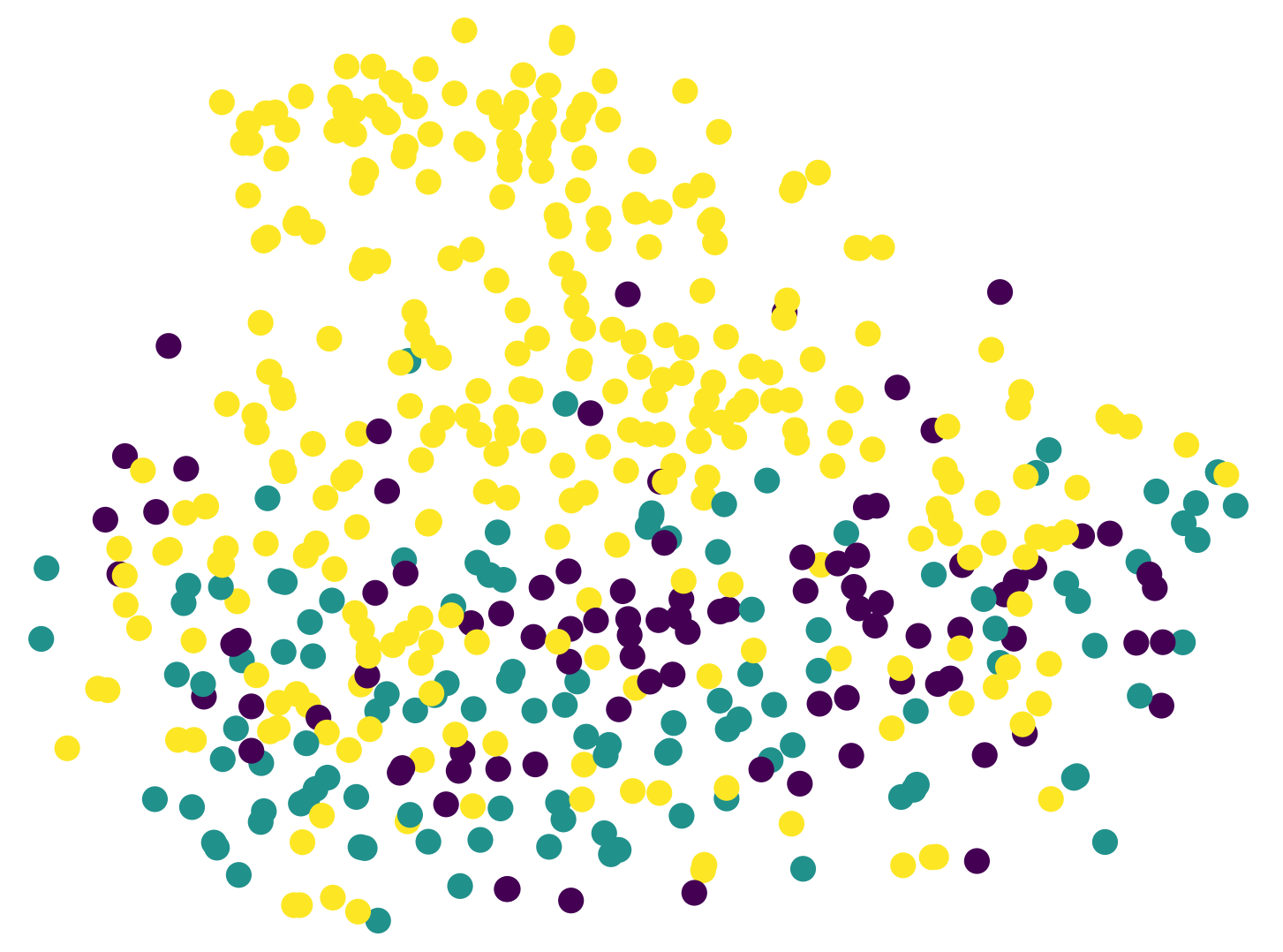}}
    \subfigure[w/o BERT]{
        \includegraphics[width=0.45\columnwidth]{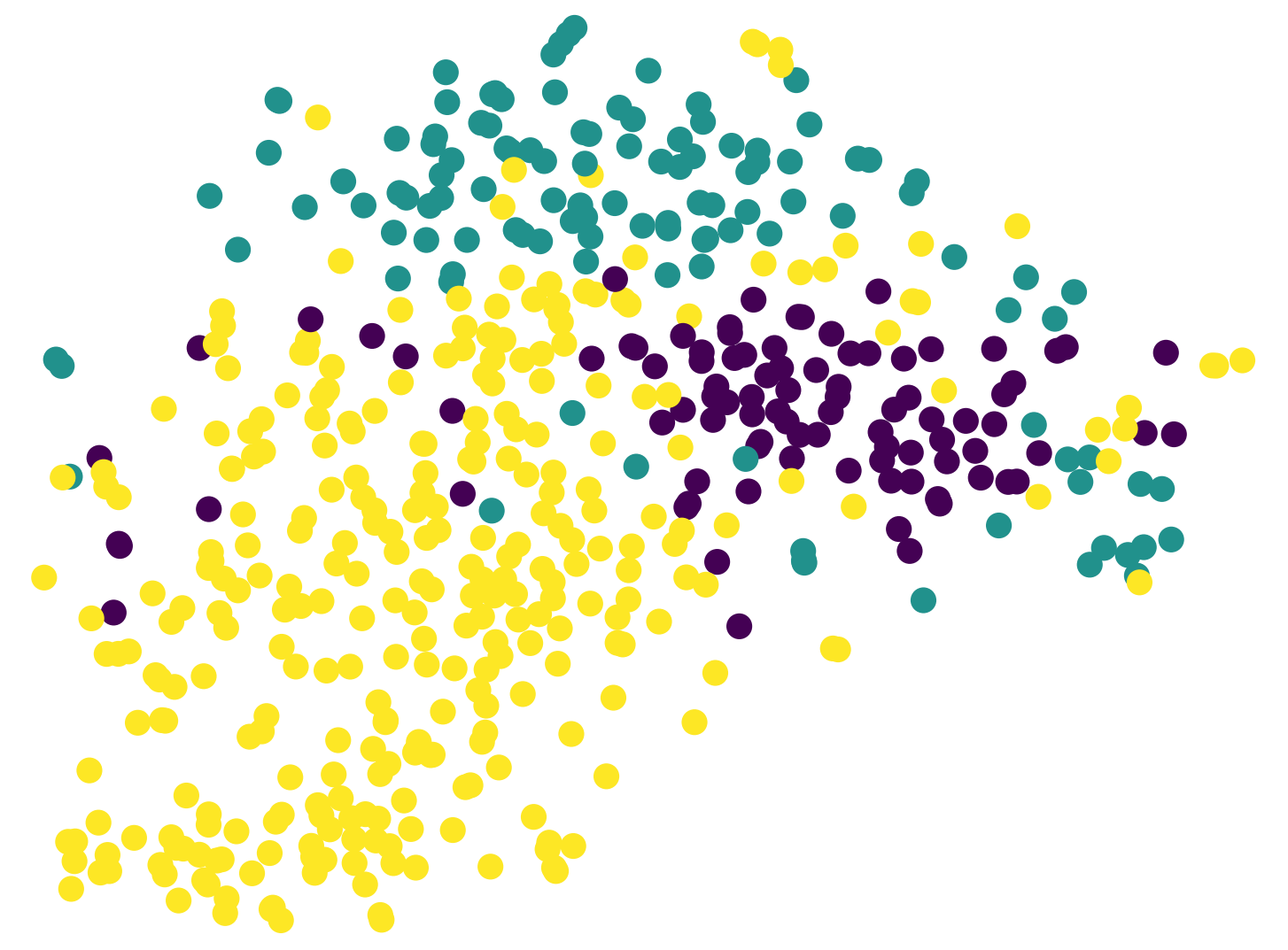}}
    \subfigure[w/o LDA]{
        \includegraphics[width=0.45\columnwidth]{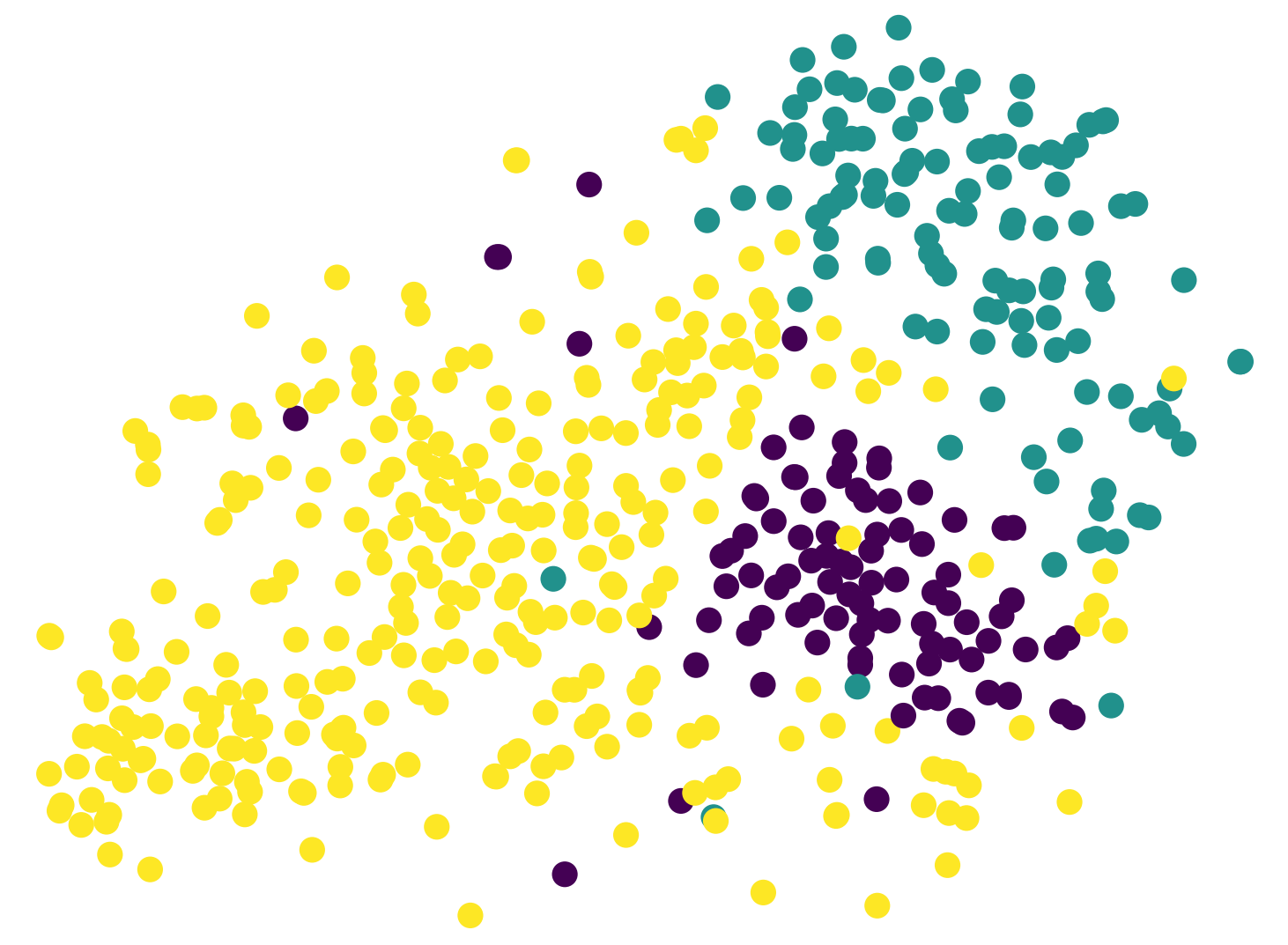}}
    \subfigure[CoSD]{
        \includegraphics[width=0.45\columnwidth]{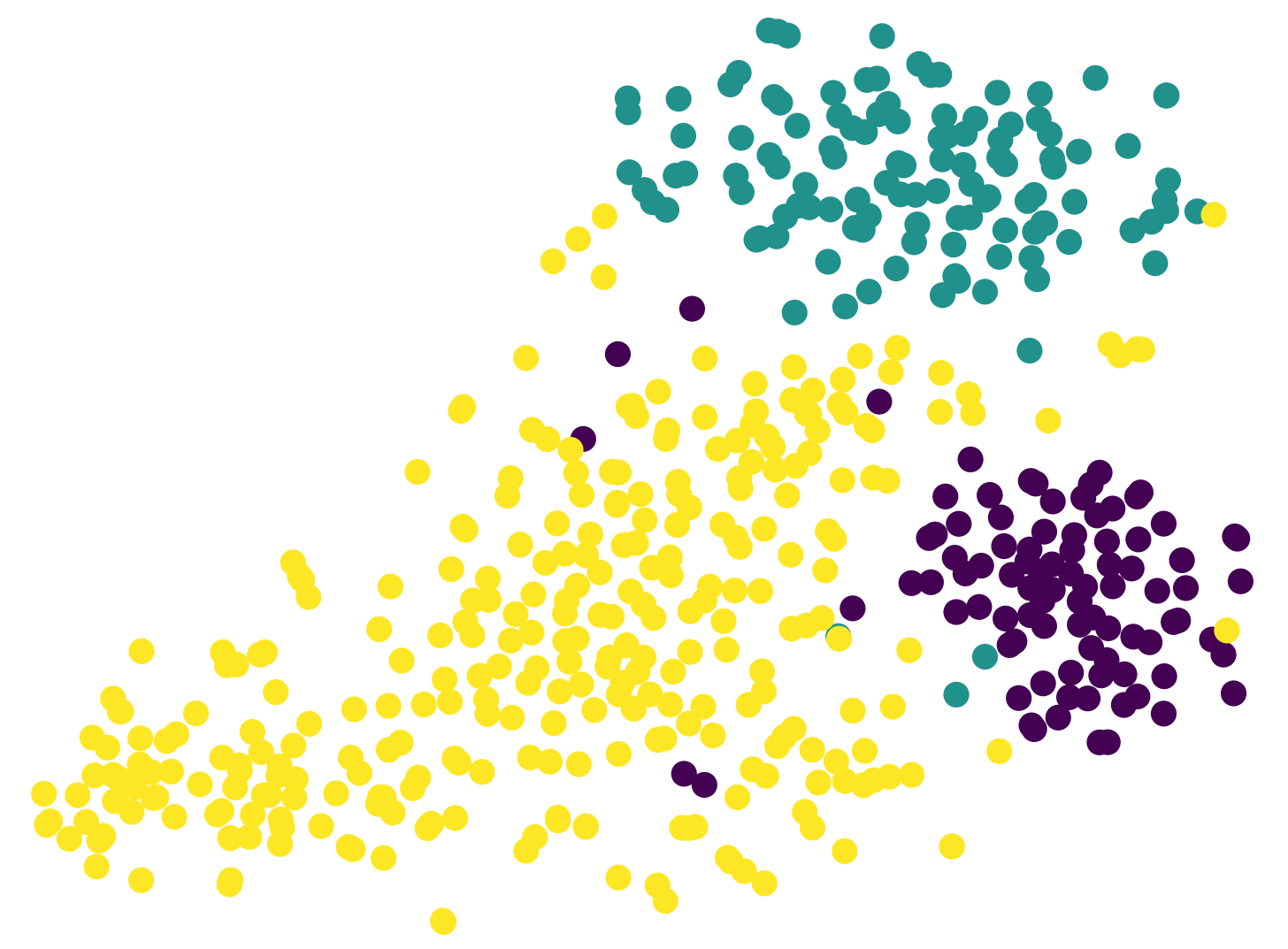}}
        \vspace{-2mm}
	\caption{T-SNE visualization of Target 'MW' on UKP}
    \label{fig:t-sne2}
\end{figure}

\begin{figure}[!t]
	\centering
    \subfigure[initial]{
        \includegraphics[width=0.45\columnwidth]{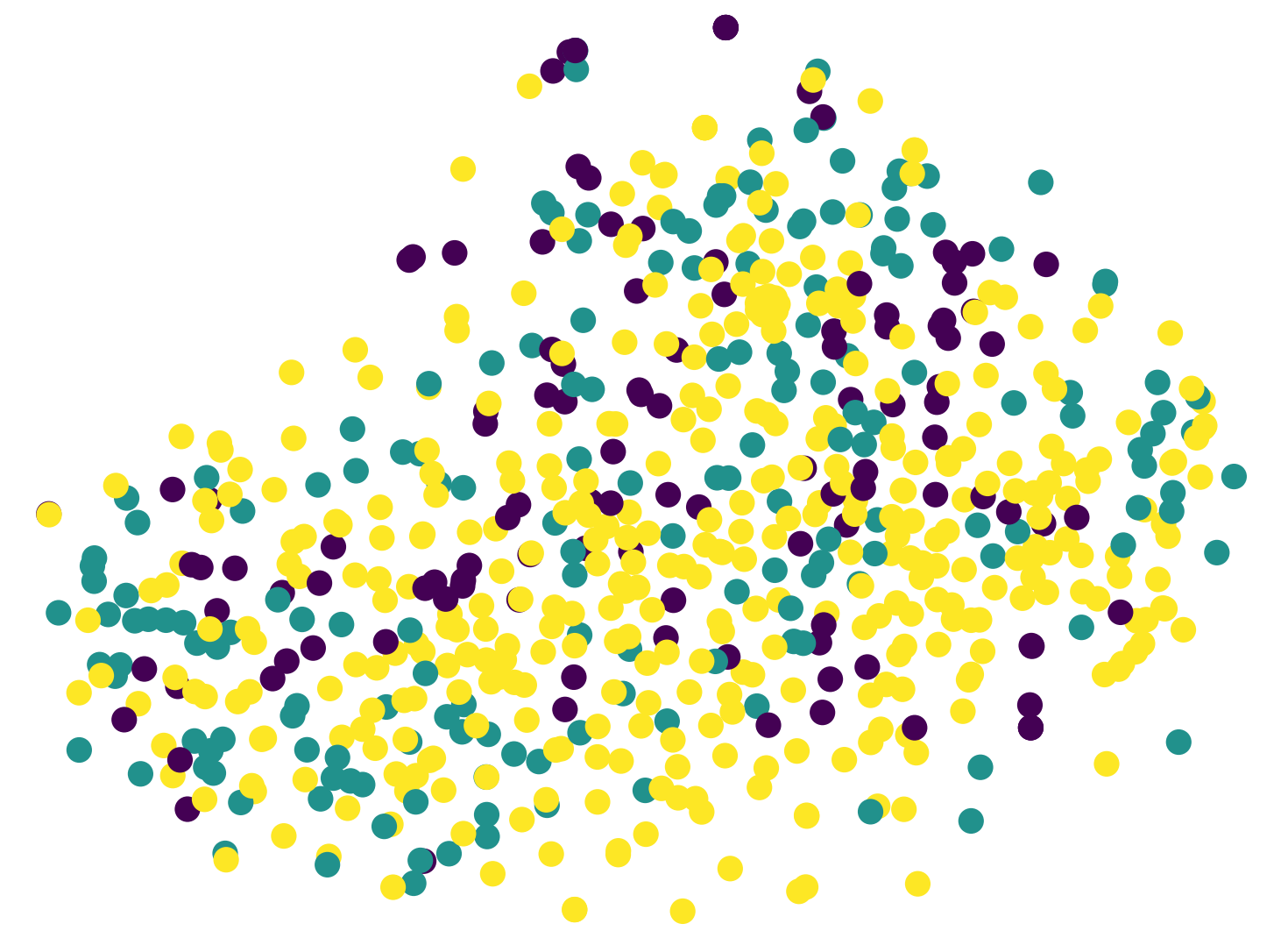}}
    \subfigure[w/o BERT]{
        \includegraphics[width=0.45\columnwidth]{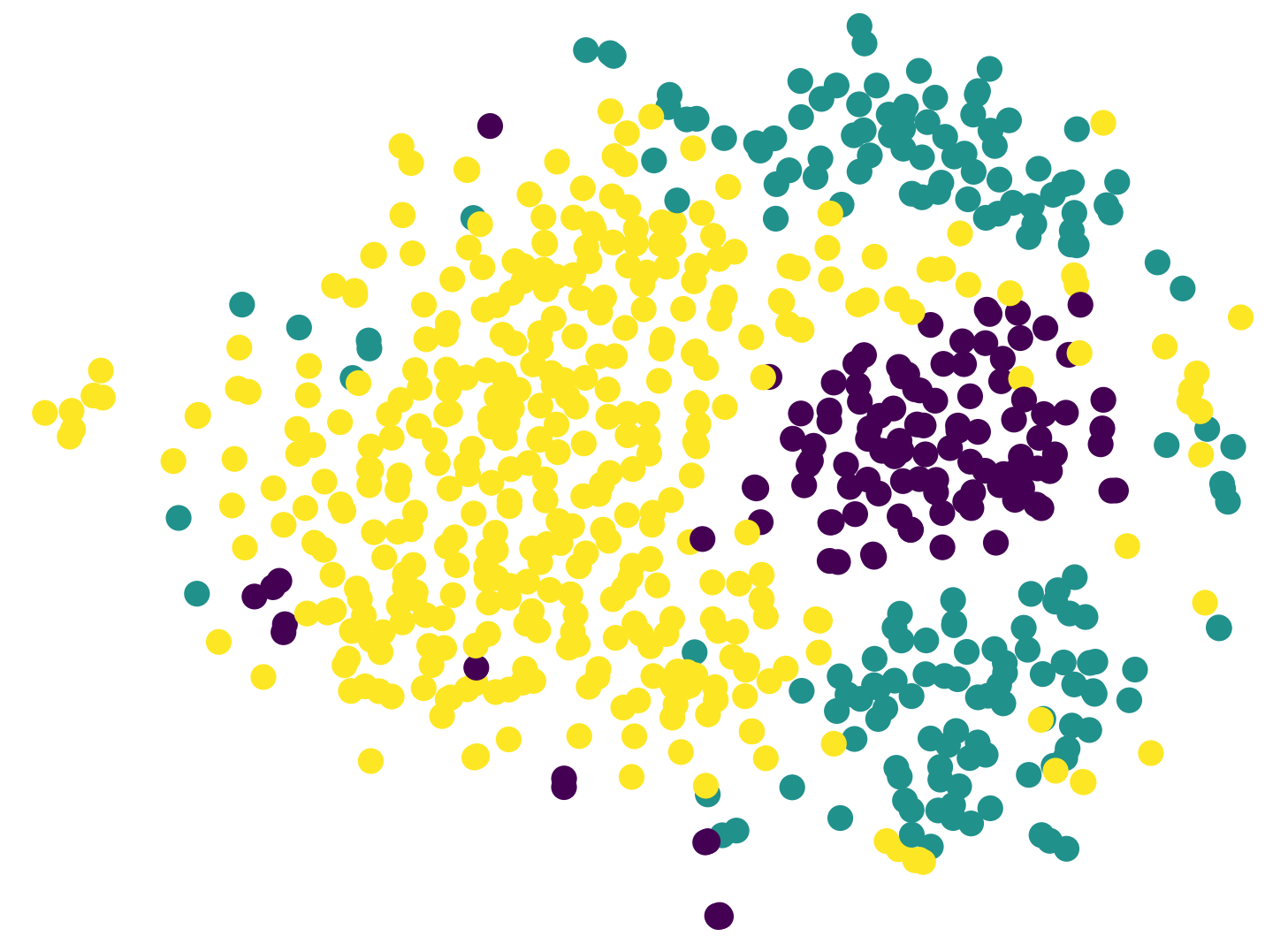}}
    \subfigure[w/o LDA]{
        \includegraphics[width=0.45\columnwidth]{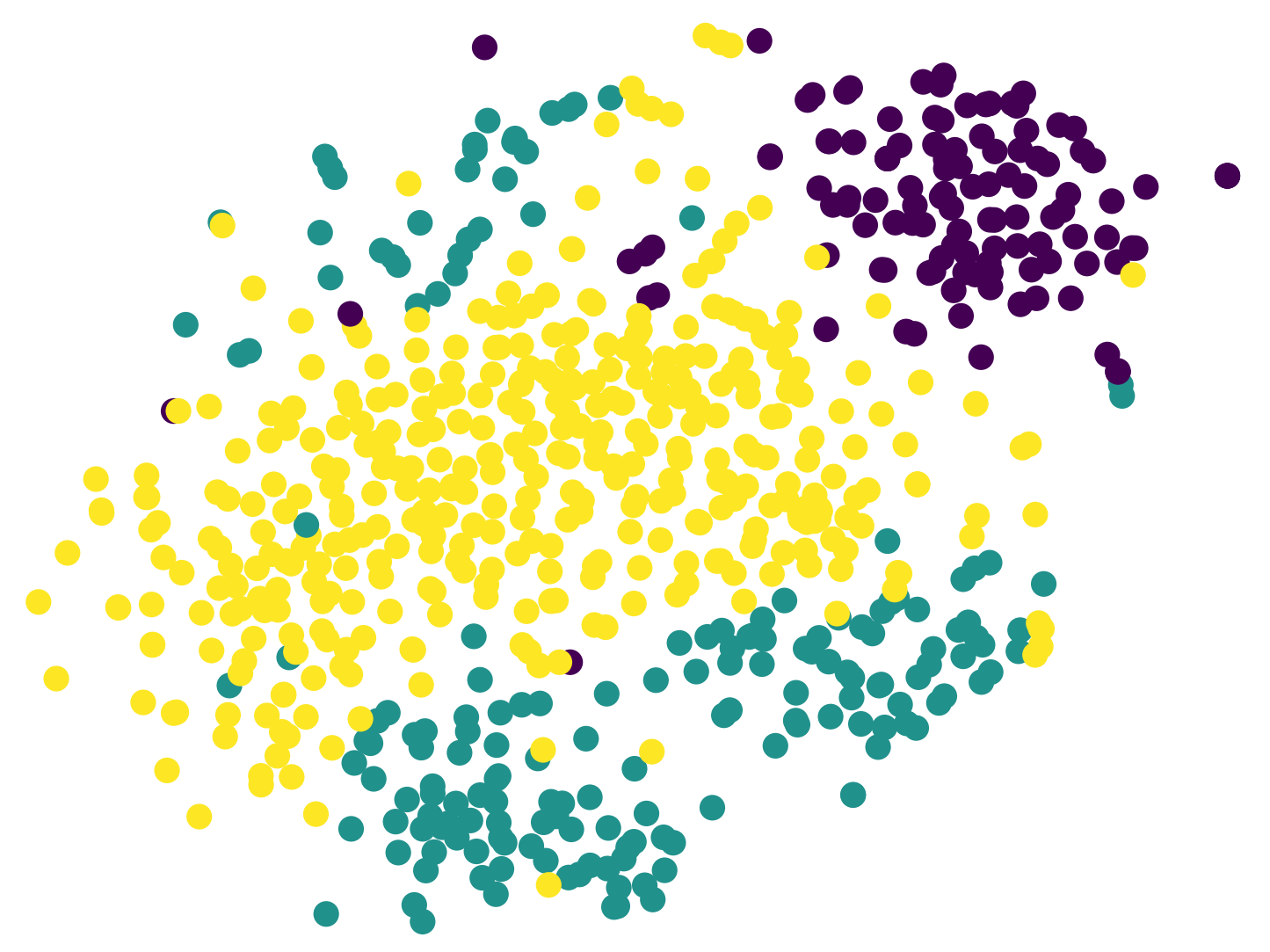}}
    \subfigure[CoSD]{
        \includegraphics[width=0.45\columnwidth]{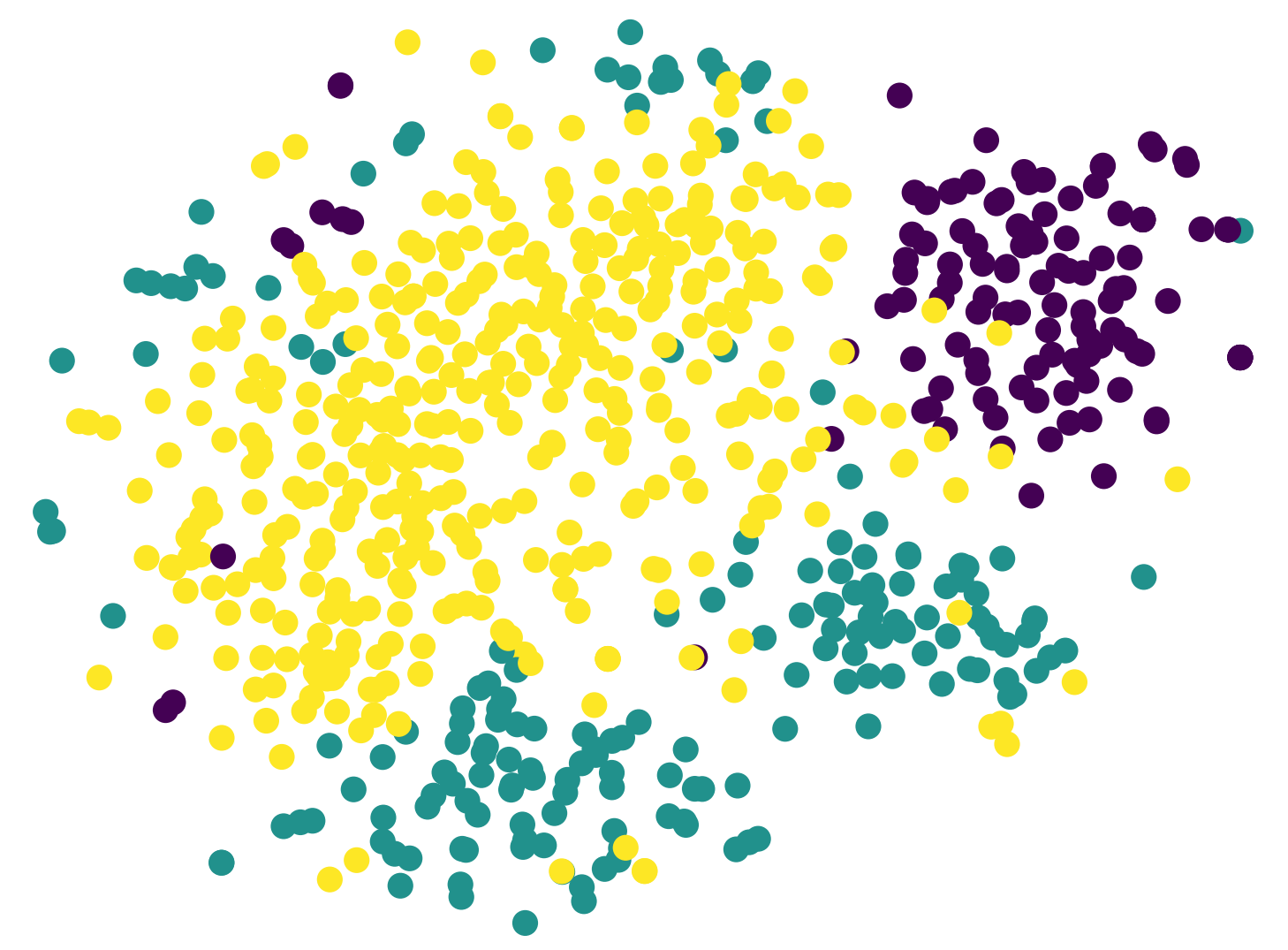}}
        \vspace{-2mm}
	\caption{T-SNE visualization of the target 'NE' on UKP}
    \label{fig:t-sne3}
\end{figure}

We selected target 'HC' on SemEval-2016, 'MW' and 'NE' on UKP, and the T-SNE visualization results of their features are presented in Figure \ref{fig:t-sne1}, \ref{fig:t-sne2} and \ref{fig:t-sne3}, where w/o BERT represents \textbf{CoSD $\neg$ BERT} and w/o LDA represents \textbf{CoSD $\neg$ LDA}.  Initially, the representations appear scattered, but after training, they are well-clustered. Three distinct clusters emerge, corresponding to the three different polarities of the stance detection task. 

In the results without BERT, the cross-range between clusters is large, and it's observed that it is divided into three clusters in the absence of LDA. This observation is attributed to the clearer semantic features associated with these targets, which also verifies the effectiveness of adding node semantic information, because our collaborative learning framework focuses more on the collaborative signaling connections between nodes, as for the nodes themselves there is a certain degree of lack of consideration of semantic information. Our method CoSD achieves the best separation effect, with a significant distance between clusters, consistent with the results of our ablation experiments in \ref{ablation}. 

\subsection{Case Study} \label{case}
To overcome the limitation of unimodality in common classification models, our new framework provides each text and stance with a specific scenario to derive their optimal representation. CoSD aims to leverage collaborative signals between texts and implicit topics during the collaborative training stage and subsequently utilize these acquired signals to identify the most similar texts to assist decision-making during the hybrid inference stage. To evaluate the effectiveness of our model and the utilization of collaborative signals, Table \ref{tab:7} illustrates several correctly classified cases, for each test text, we display the top 2 texts with the most similar representations. Similar content is highlighted in green (Favor samples) or red (Against samples) fonts. The presence of significant similarities indicates that CoSD can effectively capture collaborative signals and leverage them to identify similar texts to aid in making accurate classification decisions.

In the first example, the target is "AB" (Abortion). Comparing the test text $Test$ with the two texts $T_1$ and $T_2$ that are most similar to it, we observe that, in addition to the shared target word "abortion", phrases "Constitutional rights" in $Test$ and "rights in constitutional politics", "constitutional law" in $T_1$, as well as "U.S. Constitution" in $T_2$ can be seen as reflecting the implicit topic of "Legislation and Law". Furthermore, "for all Americans" in $Test$ and "Equal Protection Clause" in $T_1$ and "traditional sex-role stereotypes", "equality argument" in $T_2$ can be associated with the implicit topic of "Equity and Justice". Through the collaborative learning framework, we identify samples similar to $Test$ that provide valuable references, as they share the same implicit topics. Combined with Figure \ref{fig:problem}, we observe that the framework can effectively filter out texts with the most similar implicit topic distribution (not $T_2$ or $T_3$ in Figure \ref{fig:problem}).

Another example at "ML" (Marijuana Legalization). Apart from the shared target word "marijuana", phrases such as "expanded use" in $Test$ and "Increased drug use" in $T_1$ and $T_2$ can be seen as reflecting the implicit topic of "Usage Amount". Similarly, expressions like "detrimental to our youth, to public health, and to the safety of our society" in $Test$ and "negative health effects, and negative effects on families" in $T_1$, as well as "negative health effects" in $T_2$, align with the implicit topic of "Negative Impacts".

Through such a case study, it can be observed that our model can provide a classification basis for target texts from the perspective of implicit topics by characterizing similar texts. This feature enhances the explainability of the model's classification results.


\begin{figure*}[t] 
    \centering
    
    \subfigure["CC" on SemEval-2016]{
    \begin{minipage}[b]{0.49\textwidth} 
        \centering
        \includegraphics[width=\textwidth]{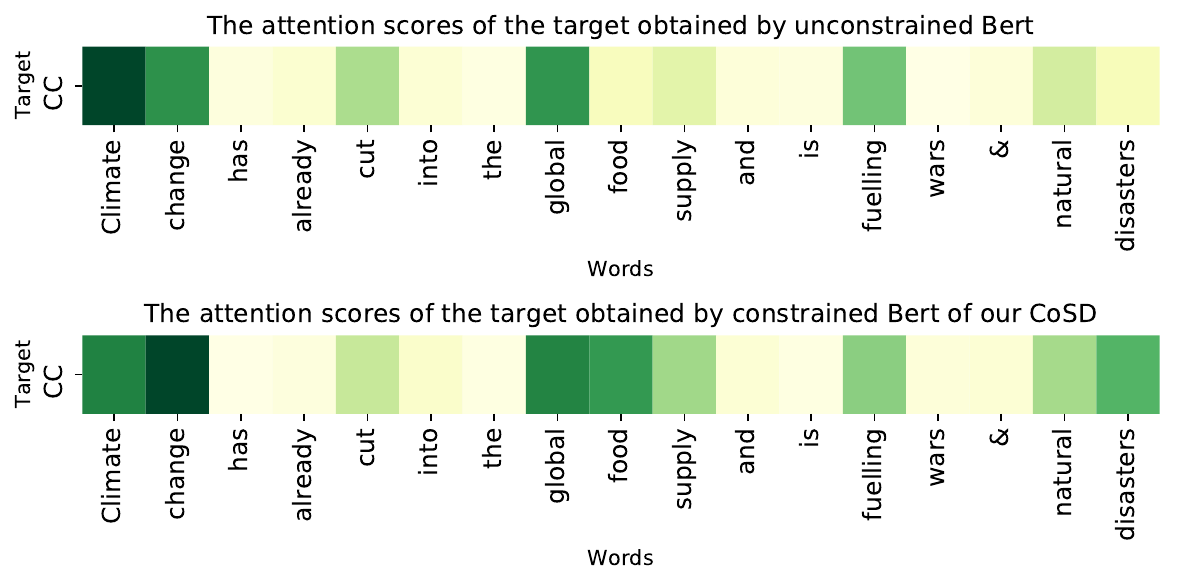}
    \end{minipage}}
    \hfill 
    \subfigure["AB" on UKP]{
    \begin{minipage}[b]{0.49\textwidth} 
        \centering
        \includegraphics[width=\textwidth]{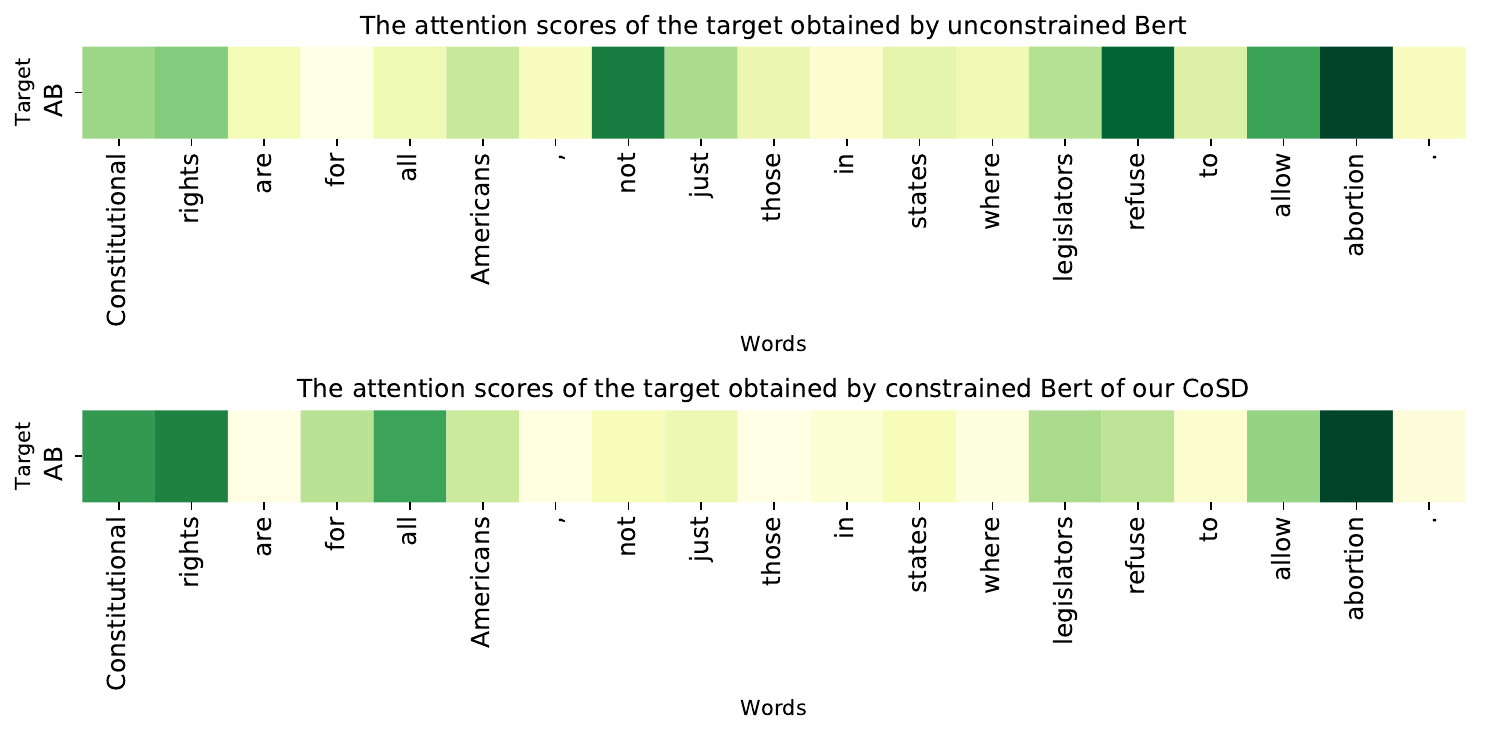}
    \end{minipage}}
    \vspace{0.05cm} 
    \subfigure["ML" on UKP]{
    \begin{minipage}[b]{0.49\textwidth} 
        \centering
        \includegraphics[width=\textwidth]{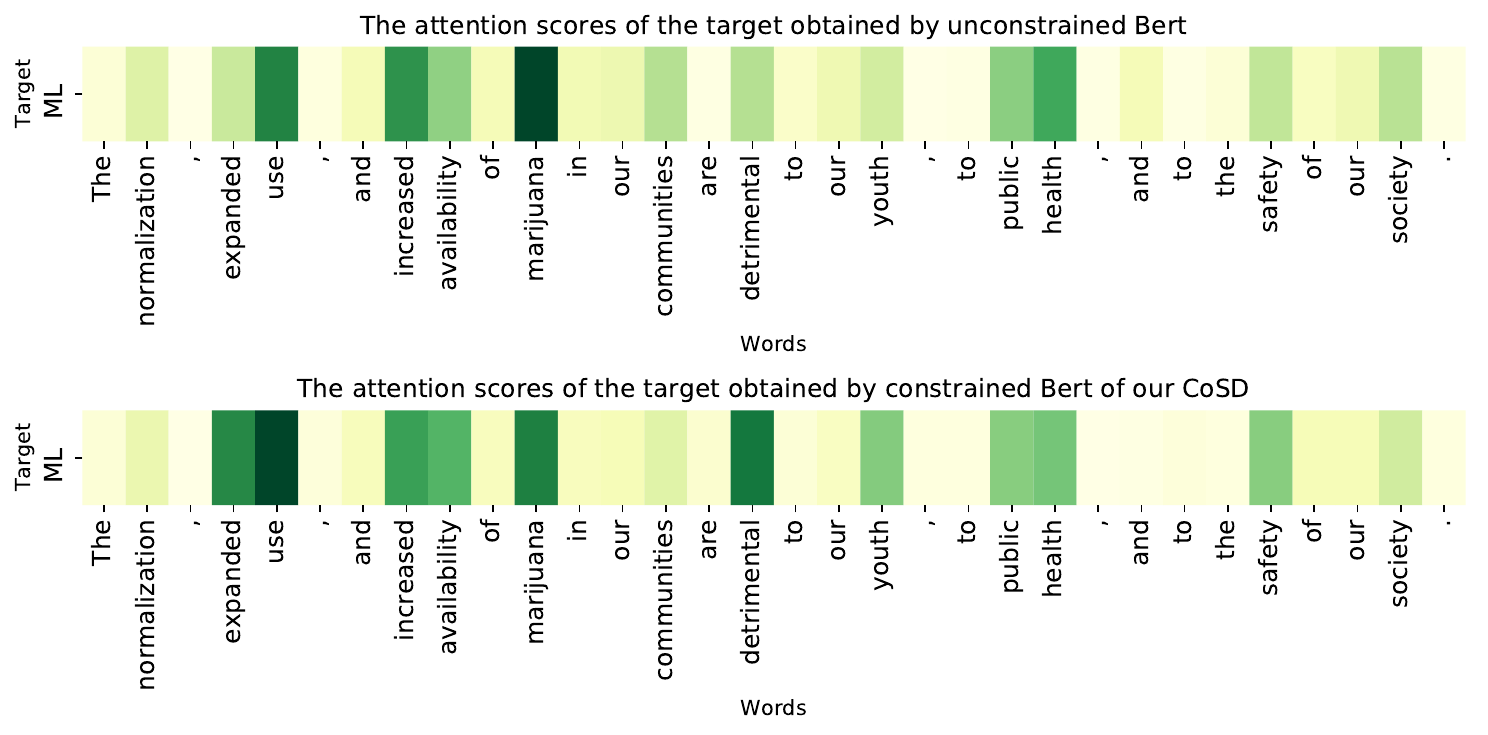}
    \end{minipage}}
    \hfill 
    \subfigure["SU" on UKP]{
    \begin{minipage}[b]{0.49\textwidth} 
        \centering
        \includegraphics[width=\textwidth]{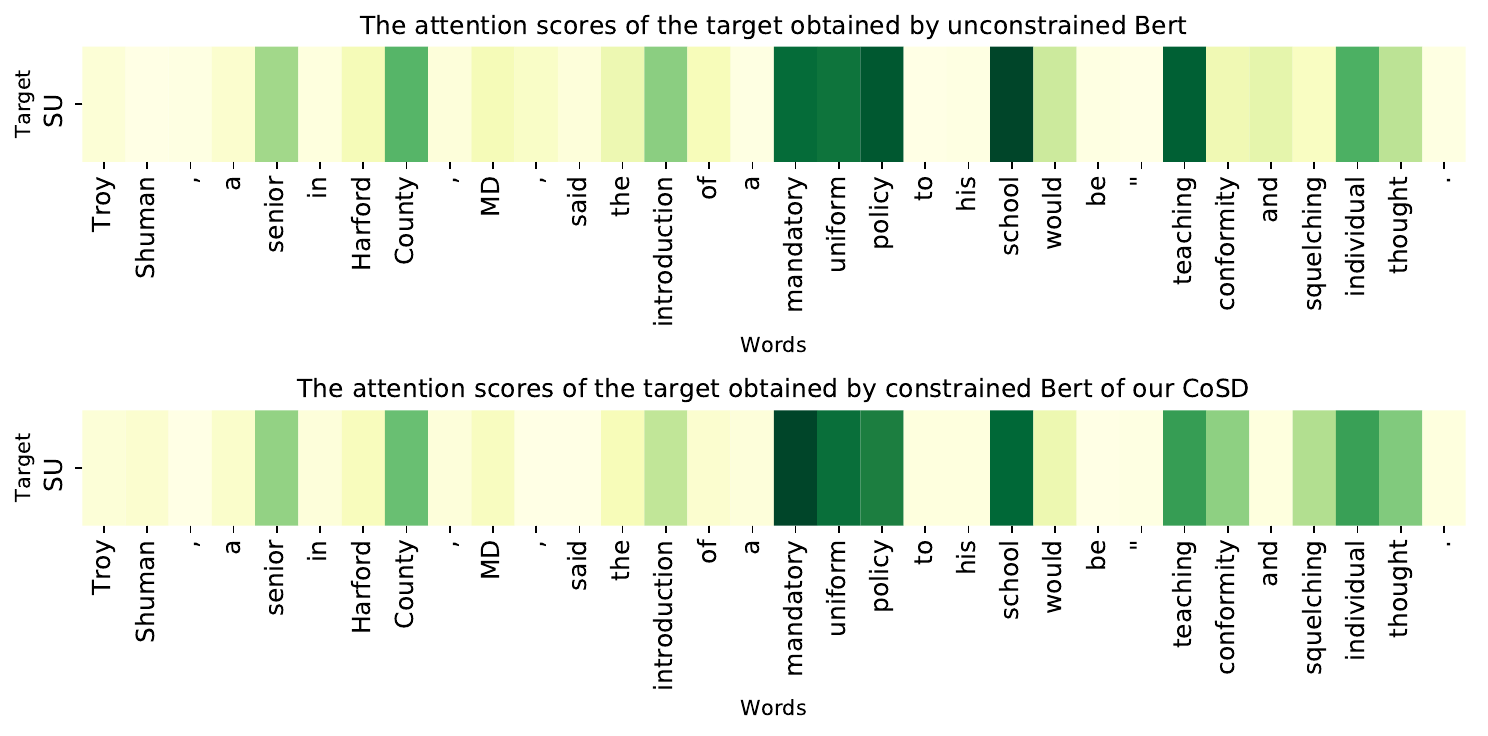}
    \end{minipage}}

	\caption{Case visualization: Visualization of attention scores for representations of BERT outputs which extract semantic information. Our constrained BERT notices terms carrying certain implicit topic information, which provides some explainability.}
    \label{fig:case visualization}
\end{figure*}

\subsection{Case Visualization} \label{case visualization}
During training, the normalized cosine similarity loss employs the embedding representation table to constrain the semantic representation of Bert's output and inject some implicit topic information into it. We visualize the attention score in Equation \ref{con:equation13} and compare it with the unconstrained BERT semantic representation, as shown in Figure \ref{fig:case visualization}. In general, after constraint, BERT tends to allocate more attention to words related to the "implicit topics". As shown in Figure \ref{fig:case visualization} (a), we observe that conventional BERT primarily focuses on "global" and "fuelling" when "climate change" is mentioned, likely associating it with concepts "global warming" and "fossil fuels". However, in this context, these associations are not accurate. Our constrained BERT identifies the "global food supply" as a "natural disaster". In Figure \ref{fig:case visualization} (b), unconstrained BERT captures words with obvious tendencies like "not", "refuse" and "allow", which may lead to confusing information. In contrast, our constrained BERT notices terms such as "Constitutional rights" and "for all Americans", which carry implicit topic information and provide explainability. 

Figure \ref{fig:case visualization} (c), (d) respectively depict a case related to target "ML" and "SU" from the UKP dataset. While constrained BERT does pay more attention to the expression of some continuous implicit topic words, the differences in the focus between constrained and unconstrained BERT are not particularly significant. We infer that this may also explain the modest improvement between CoSD $\neg$ LDA and CoSD in the \ref{ablation}. Through our framework, BERT is injected with a certain amount of implicit topic information obtained by collaborative training. This enhancement empowers BERT to play a more influential role. Consequently, in \ref{visualization}, the performance of TSE w/o Bert is slightly inferior to that w/o LDA.

\section{Conclusion}
In this paper, we propose a new stance detection framework, CoSD, which establishes indirect connections between texts and targets by constructing a Heterogeneous Topic Graph. Then, the Collaboration Propagation Aggregation Module is leveraged to capture the collaboration signals between the text and the target. In the end, the Inference Stage yields the correct classification of the text. The experimental results demonstrate that among the state-of-the-art methods, our proposed method CoSD achieves the SOTA performance. For future work, we aim to further explore implicit information shared within the text for zero-shot stance detection.

\section*{Acknowledgments}
This work is supported by the National Natural Science Foundation of China (No. 62372043) and by the BIT  Research and  Innovation Promoting Project (Grant No.2023YCXY037). This work is also supported by the Shanghai Baiyulan Talent Plan Pujiang Project (23PJ1413800).


\bibliographystyle{IEEEtran}
\bibliography{sample-base}

\begin{IEEEbiography}[{\includegraphics[width=1in,height=1.25in,clip,keepaspectratio]{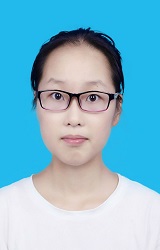}}]{Yinghan Cheng}
	is currently studying for her master’s degree at Beijing Institute of Technology China in Computer Science. Her research interests focus on stance detection, sarcasm
	detection, and sentiment analysis.
\end{IEEEbiography}

\begin{IEEEbiography}[{\includegraphics[width=1in,height=1.25in,clip,keepaspectratio]{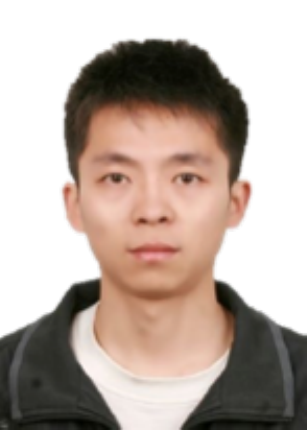}}]{Qi Zhang}
	received the dual PhD degree from the Beijing Institute of Technology, Beijing, China, and the University of Technology Sydney, Sydney, NSW, Australia. 
	He is currently a Research Fellow at Tongji University, Shanghai, China. He has authored high-quality papers in premier conferences and journals, including NeurIPS, AAAI, IJCAI, TheWebConf, TKDE, TNNLS, and TOIS. His primary research interests include collaborative filtering, recommendation, learning to hash, and MTS analysis.
\end{IEEEbiography}
\vspace{-5mm}
\begin{IEEEbiography}[{\includegraphics[width=1in,height=1.25in,clip,keepaspectratio]{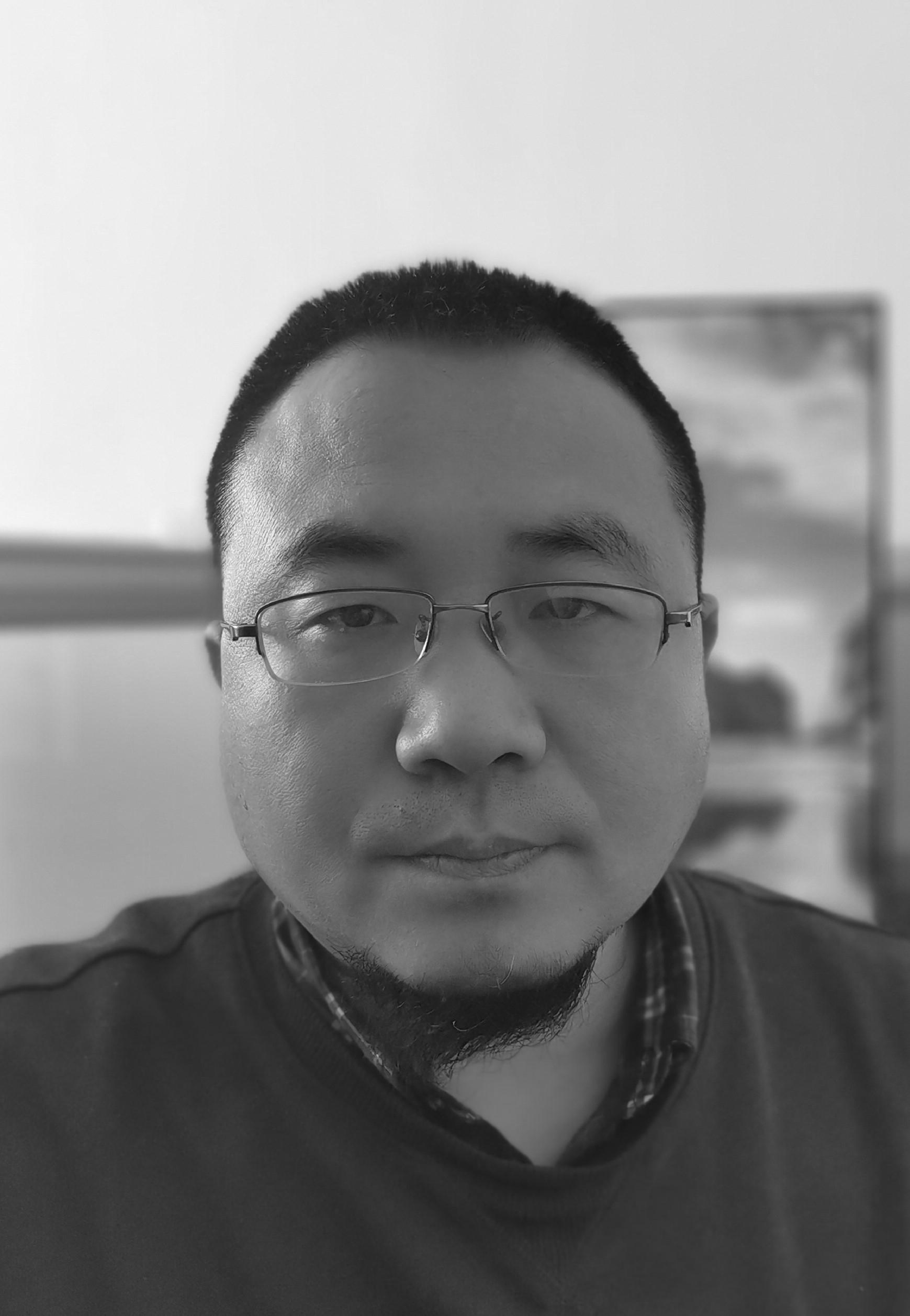}}]{Chongyang Shi} is currently an associate professor at the School of Computer Science, Beijing Institute of Technology. He received his Ph.D. degree from BIT in 2010 in computer science. Dr. Shi's research areas focus on information retrieval, knowledge engineering, personalized service, sentiment analysis, etc. He serves as an editorial board member for several international journals and has published more than 40 papers in international journals and conferences.
\end{IEEEbiography}
\vspace{-5mm}
\begin{IEEEbiography}[{\includegraphics[width=1in,height=1.25in,clip,keepaspectratio]{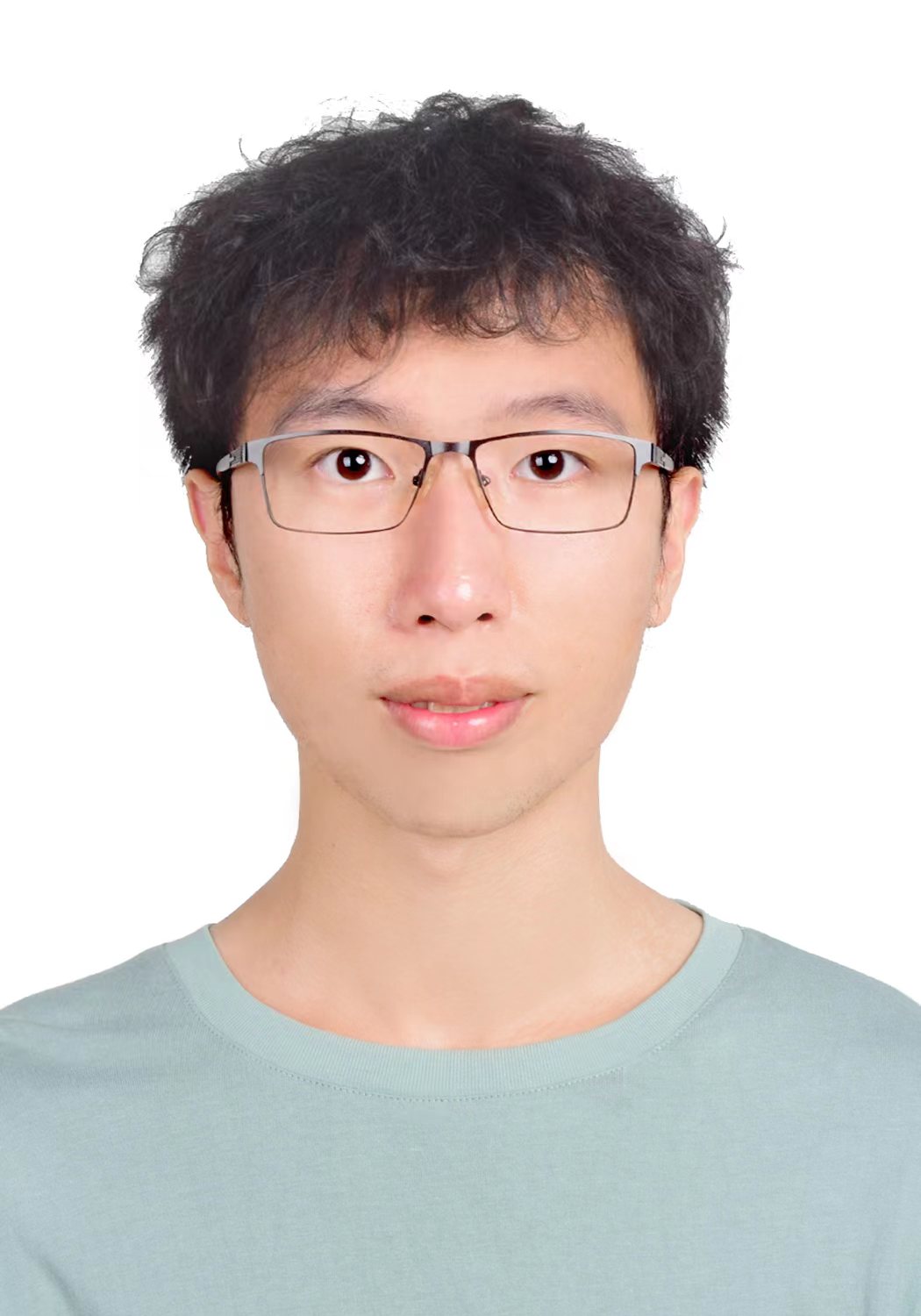}}]{Liang Xiao} is currently studying for his master's degree in Beijing lnstitute of Technology, Beijing, China in Computer Science. His research interests focus on clickbait detection, fake news detection, and large language models.
\end{IEEEbiography}
\vspace{-5mm}
\begin{IEEEbiography}[{\includegraphics[width=1in,height=1.25in,clip,keepaspectratio]{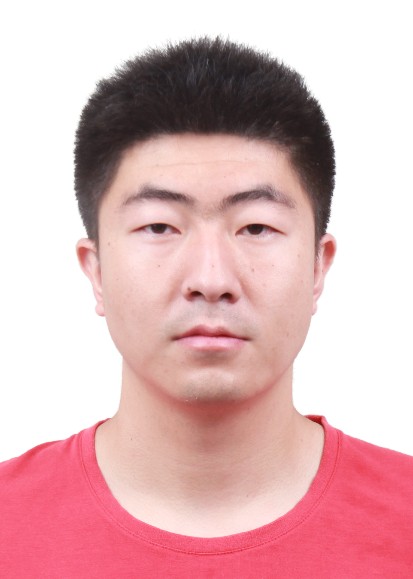}}]{Shufeng Hao} received the Ph.D. degree from Beijing Institute of Technology, China. He is currently a lecturer at Taiyuan University of Technology. His research interests include sentiment analysis, document retrieval, and deep learning.
\end{IEEEbiography}
\vspace{-5mm}
\begin{IEEEbiography}[{\includegraphics[width=1in,height=1.25in,clip,keepaspectratio]{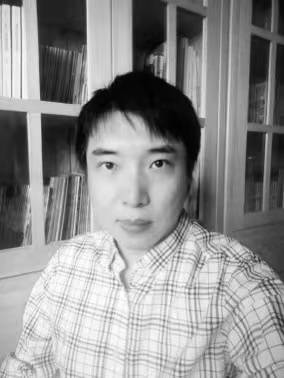}}]{Liang Hu} received dual Ph.D. degrees from Shanghai Jiao Tong University, China and University of Technology Sydney, Australia. He is currently a distinguished research fellow at Tongji University and chief AI scientist at DeepBlue Academy of Sciences. His research interests include recommender systems, machine learning, data science, and general intelligence. 
\end{IEEEbiography}

\end{document}